\documentclass[preprint]{elsarticle}

\usepackage[utf8]{inputenc}
\usepackage{xcolor}
\usepackage{lineno}
\usepackage[title]{appendix}

\usepackage[letterpaper,top=2cm,bottom=2cm,left=2cm,right=2cm,marginparwidth=1.75cm]{geometry}

\usepackage{amsmath}
\usepackage{palatino,graphicx,wrapfig}
\usepackage[small,it]{caption}
\usepackage{amsmath}
\usepackage{amssymb,bm}
\usepackage{parskip}
\usepackage{multirow}

\journal{}

\title{{Interpretable A-posteriori Error Indication for Graph Neural Network Surrogate Models}}

\author[label1]{Shivam Barwey}
\author[label3]{Hojin Kim}
\author[label2,label3]{Romit Maulik}

\affiliation[label1]{organization={Transportation and Power Systems Division, Argonne National Laboratory},
            city={Lemont},
            postcode={60439},
            state={IL},
            country={USA}}

\affiliation[label2]{organization={Mathematics and Computer Science Division, Argonne National Laboratory},
            city={Lemont},
            postcode={60439},
            state={IL},
            country={USA}}

\affiliation[label3]{organization={College of Information Sciences and Technology, Pennsylvania State University},
            city={University Park},
            postcode={16802},
            state={PA},
            country={USA}}

\date{June 2024}

\begin{document}

\begin{abstract}
Data-driven surrogate modeling has surged in capability in recent years with the emergence of graph neural networks (GNNs), which can operate directly on mesh-based representations of data. The goal of this work is to introduce an interpretability {enhancement} procedure for GNNs, with application to unstructured mesh-based fluid dynamics modeling. {Given a black-box baseline GNN model}, the end result is an {interpretable GNN} model that isolates regions in physical space, corresponding to sub-graphs, that are intrinsically linked to the forecasting task while retaining the predictive capability of the baseline. These structures identified by the {interpretable GNN}s are adaptively produced in the forward pass and serve as explainable links between the baseline model architecture, the optimization goal, and known problem-specific physics. Additionally, through a regularization procedure, the {interpretable GNN}s can also be used to identify, during inference, graph nodes that correspond to a majority of the anticipated forecasting error, adding a novel interpretable error-tagging capability to baseline models. Demonstrations are performed using unstructured flow field data sourced from flow over a backward-facing step at high Reynolds numbers, with geometry extrapolations demonstrated for ramp and wall-mounted cube configurations.
\end{abstract}


\maketitle
\tableofcontents

\section{Introduction}
Graph neural networks (GNNs) have gained immense popularity in the scientific machine learning community due to their ability to learn from graph-based representations of data \cite{wu2022graph,geometric_dl}. In this context, graph-based data is represented as a collection of nodes and edges. The set of edges denotes connections between nodes based on a predefined distance measure in a node feature space, constituting the connectivity or adjacency matrix of the graph. GNNs utilize this connectivity to create complex non-local models for information exchange among node features in their respective neighborhoods. The definition of this graph connectivity implicitly invokes an inductive bias in the neural network formulation based on the way in which the connectivity is constructed, which can lead to significant benefits when it comes to model accuracy and generalization \cite{battaglia_2018}. {As such, GNNs have achieved state-of-the-art results for several notoriously difficult physical modeling problems, including protein folding \cite{alphafold}, solving partial differential equations \cite{gao2022physics}, and data-based weather prediction \cite{graphcast}}. 

Although conventional neural network architectures (e.g., convolutional neural networks \cite{romit_pof_2021,morimoto2021convolutional} and multi-layer perceptrons \cite{romit_jfm_2017,eivazi2020deep}) have seen success in various fluid dynamics modeling applications, GNNs offer an appealing advantage, in that they can readily interface with mesh-based and unstructured representations of data \cite{meshgraphnets,gilmer_2017,battaglia_2018}.
In other words, since graph connectivity can itself be equivalent to a spatial mesh on which the system solution is discretized, GNNs offer a natural data-based modeling framework for fluid flow simulations. This, in turn, translates into (a) an ability to model flow evolution in complex geometries described by unstructured meshes, and (b) an ability to extrapolate models to unseen geometries, both of which are significant advantages in practical engineering applications. These qualities have set new standards for data-based surrogate modeling in both steady \cite{belbute2020combining,yang2022amgnet} and unsteady \cite{lino_gnn,multiscale_meshgraphnets} fluid flow applications. Additionally, since GNNs in this setting share a similar starting point to standard numerical solution procedures (a mesh-based domain discretization), GNN message passing layers share considerable overlap with vetted solution methods used in the computational fluid dynamics (CFD) community, such as finite difference, finite volume, and finite element methods -- this quality has been used to improve baseline GNN architectures to develop more robust and stable surrogate models in more complex scenarios \cite{karthik_gnn,physgnn}, establishing a promising middle-ground between purely physics-based and data-based simulation strategies. 

\begin{figure}
    \centering
    \includegraphics[width=\textwidth]{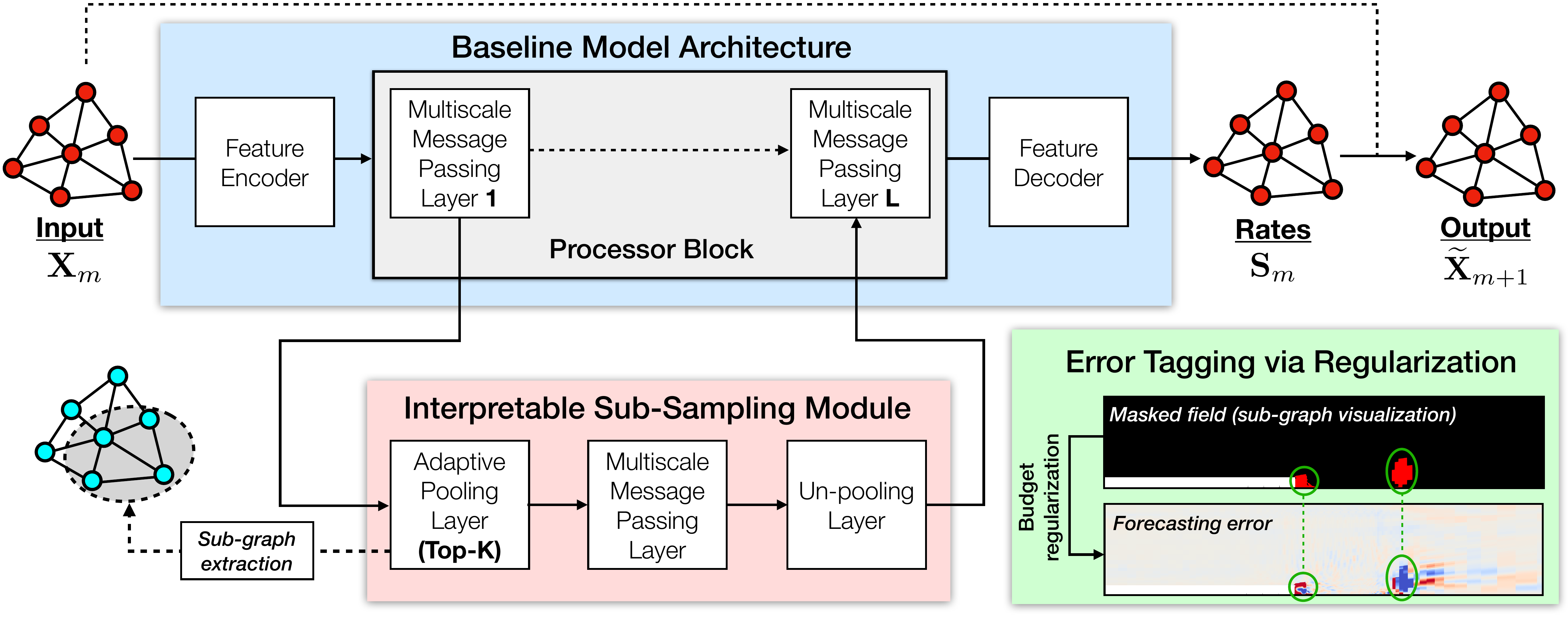}
    \caption{\textbf{(Top, blue box)} Schematic of baseline GNN model architecture. \textbf{(Bottom, red box)} Schematic of interpretable {sub-sampling} module. Module is implemented via skip connection between first and last message passing layers in baseline processor block. Interpretability comes from adaptive pooling layer, which generates sub-graphs {via sub-sampled nodes} that can be visualized as masked fields. Using a regularization strategy, masked fields can then be used to identify regions of forecasting approximation errors during model evaluation. Details provided in Sec.~\ref{sec:results:fine_tuning}. \textbf{(Bottom, green box)} Error tagging procedure by means of a budget regularization strategy that encourages prediction errors to be contained within the extracted nodes in sub-graphs (the masked fields). Details provided in Sec.~\ref{sec:results:budget}.}
    \label{fig:arch}
\end{figure}

The goal of this work is to introduce additional, equally critical advantages that can be provided within the GNN framework: (1) the ability to generate interpretable latent spaces (or latent graphs) as a by-product of the network architecture through {the addition of a graph sub-sampling module}, and (2) the ability to use these interpretable latent graphs to adaptively identify, or tag, sub-graphs that contribute most to the prediction error (i.e., a-posteriori error tagging). To emphasize broad applicability, these goals are achieved here in the context of {interpretability enhancements to baseline models}: more specifically, when given a pre-trained baseline GNN surrogate model, this work shows how appending an additional graph-based trainable module can not only enhance the interpretability properties of the baseline in various modeling tasks (with focus on mesh-based fluid flow forecasting), but also provide key modeling benefits in the form of error tagging. This strategy is illustrated in Fig.~\ref{fig:arch}.

Before outlining additional contributions of the current study in greater detail, it is first important to emphasize the increasing demand for interpretable (or explainable) artificial intelligence \cite{xai}. Although interpretability has been a core property of traditional data-driven models for fluid flows -- such as proper orthogonal decomposition \cite{berkooz_pod} and other modal decomposition approaches \cite{mezic_arfm,schmid_arfm,kaiser_jfm_2014} relying on linear projection operations for dimensionality reduction -- the ability to better interpret models built around more expressive nonlinear neural network architectures is becoming a highly valuable asset. This is particularly important in engineering applications, where access to failure modes and model-form errors can prevent potentially high-cost catastrophic device or system operation failures \cite{rudin_2019}.

Interpretability in neural networks can be categorized into two branches: passive approaches and active approaches \cite{nn_interp_review}. Passive interpretability occurs after the training stage, and utilizes post-processing tools (e.g., sensitivity analysis, parameter decompositions and visualization) to better understand the input-to-output mapping process. This is a particularly useful strategy for probing the decision-making process in very large neural networks \cite{samek2016evaluating}, or for identifying causal relationships in physics applications \cite{shivam_proci}. On the other hand, active interpretability occurs before the training stage and involves architectural or objective function modifications. For example, interpretability-enhancing additions to standard objective functions have been used to train a convolutional neural network such that regions of influence of each hidden layer in the model could be readily accessed \cite{interpretable_cnn}. In physics applications, particularly for rapid forecasting of fluid dynamics, interpretability properties can be baked into neural network approaches in many different problem-specific ways. Physics-informed neural networks leverage known partial differential equations (PDEs) to augment standard supervised training objective functions, which in turn loosely constrains the neural network forward pass to adhere to existing physical models \cite{raissi_pinn}. Other methods (e.g., those based on operator learning or basis function expansions) enhance interpretability through direct neural network layer modifications \cite{deeponet,vinuesa_interpretable} or by leveraging classification-based regression strategies \cite{shivam_ftc,romit_blending}. 

Although many of these existing strategies for interpretability based on incorporation of known physics -- such as the utilization of PDE discretization operators into layer operations and objective functions -- are readily extendable to the graph neural network paradigm \cite{karthik_gnn,haber_gnn}, recent work has shown that GNNs admit alternative promising pathways for interpretable scientific machine learning in purely data-based frameworks that are not tied to single applications \cite{shivam_topk}. More specifically, in the context of mesh-based reduced-order modeling, recent work develops a multiscale GNN-based autoencoder, where an adaptive graph pooling layer is introduced to generate interpretable latent graphs~\cite{shivam_topk}. Here, the graph pooling layer -- which serves as the mechanism for graph-based dimensionality reduction in the autoencoding application -- amounts to flow field-conditioned node sampling and sensor identification, and extracts subsets of nodes that are (a) conditioned on the input graph node features, and (b) relevant to the regression task at hand. When input graph nodes coincide with locations in physical space given by unstructured mesh-based connectivities, the adaptive node sampling mechanism (a projection-truncation operation) can be visualized in this same physical space, establishing an accessible and interpretable connection between the problem physics, the GNN architecture, and the objective function. 

The scope of this previous work~\cite{shivam_topk}, however, concerns unstructured flow field reconstruction (i.e., a GNN was used to learn an interpretable identity map). Instead, the objective of this work is to (a) extend this notion of interpretability into a predictive mesh-based modeling setting, with focus on forecasting unstructured turbulent fluid flows, and (b) use such interpretability enhancements to append black-box models with explainable error-tagging capability. As such, given an input graph-based representation of an unstructured flow field, the GNN in this work models the time-evolution of node features. For this forecasting task, the objective is to show how augmenting pre-trained baseline GNNs with trainable adaptive pooling modules results in an interpretable {modeling} framework for mesh-based predictive modeling. Ultimately, the input to this procedure is an arbitrary GNN baseline model, and the output is a modified GNN that produces interpretable latent graphs tailored to different {modeling} objectives, including error-tagging capability in the context of data-based forecasting. Although the application focus here is mesh-based fluid flow modeling, the framework is expected to extend into other applications that leverage graph neural networks. The core elements of this work are as follows:
\begin{itemize}
    \item \textbf{Multiscale GNN baseline:} A baseline GNN leveraging multiscale message passing (MMP) layers is trained to model the evolution of an unstructured turbulent fluid flow. Datasets to demonstrate the methods introduced in this work are sourced from unstructured fluid dynamics simulations of flow over a backward facing step using OpenFOAM, an open-source CFD library \cite{jasak2007openfoam}. Reynolds numbers of flow trajectories ({based on the step height of the backward-facing step configuration}) range from 26000 to 46000. 
    
    \item \textbf{Interpretable {graph sub-sampling} module: } The aforementioned graph pooling layer -- which relies on a learnable node sub-sampling procedure -- is attached to the baseline GNN, creating an augmented GNN architecture with enhanced interpretability properties. {Interpretability enhancements are then provided} by freezing the parameters of the baseline GNN and ensuring the parameters in the newly introduced module are freely trainable. {The result is a so-called \textit{interpretable GNN}.}

    \item \textbf{Error tagging: } A regularization term, representing a mean-squared error budget, is added to the objective function during the {interpretable GNN} training process. The minimization of this term ensures that the {interpretable} GNN tags, during inference, a subset of the graph nodes that are expected to contribute most significantly to the GNN forecasting error. 

    \item \textbf{Geometry extrapolation: } {interpretable GNN}s -- which were trained on the BFS geometry alone -- are evaluated on two additional geometries exhibiting different flow separation physics (a ramp and a wall-mounted cube case). The objective is to assess the level of consistency in identified features and error tagging capability by means of inference on unseen configurations.
\end{itemize}

\section{Methods}
\label{sec:modeling_task}

Mesh-based data required to train the GNNs are sourced from unstructured computational fluid dynamics (CFD) simulations in a two-dimensional backward-facing step (BFS) configuration using OpenFOAM \cite{openfoam_fv}. Simulations are performed for a range of Reynolds numbers (Re) based on the BFS step height to populate training and testing sets. Specifically, 10 training set trajectories were sampled in the equi-spaced Re range of [26214, 44569], and 10 testing set trajectories in the Re range of [27233, 45589] {(the reader is referred to Appendix~\ref{app:b} for a description of corresponding friction Reynolds numbers)}. Each trajectory is comprised of $419$ two-component velocity snapshots separated by $\Delta t = 10^{-4}s$.

Graphs are generated using a dual mesh interpretation, with each mesh-based CFD snapshot representing an undirected graph with 20540 nodes and 81494 edges. In this interpretation, graph nodes coincide with centroids of computational cells, and edges are formed by connecting neighboring cell centroids such that they intersect shared cell faces. This coincides with numerical stencils used in finite volume (FV) methods in CFD \cite{fv_book}. A summary of the configuration -- including a description of the mesh, graph extraction process, flow physics, and CFD setup -- is provided in Appendix~\ref{app:a} and \ref{app:b}.

The GNN architectures introduced in this work operate under the context of mesh-based surrogate modeling. More specifically, the modeling task is 
\begin{equation}
    \label{eq:modeling_task}
    {\cal G}_\theta({\bf X}_m, {\bf E}, E) = {\bf S}_m \approx {\bf X}_{m+1} - {\bf X}_m, 
\end{equation}
where ${\cal G}_\theta$ denotes a GNN described by parameter set $\theta$. The inputs to the GNN are (1) the initial mesh-based velocity field ${\bf X}_m \in \mathbb{R}^{|V| \times N_F}$ (a node attribute matrix, where $|V|$ is the number of graph nodes and $N_F=2$ corresponds to the two velocity components stored on each node), (2) an edge attribute matrix ${\bf E} \in \mathbb{R}^{|E| \times 2}$, where $|E|$ is the number of edges, each of which stores the relative distance vector between nodes, and (3) the set of edges $E$ (the adjacency matrix). 

The GNN prediction ${\bf S}_m$ physically represents the rate-of-change of the velocity field at a fixed GNN timestep $\Delta t$ implicitly prescribed by the training data. The model $\cal G$ can then be used to make forecasts through a residual update via
\begin{equation}
    \label{eq:gnn_prediction}
    \widetilde{{\bf X}}_{m+1} = {\cal G}_\theta({\bf X}_m, {\bf E}, E) + {\bf X}_m, 
\end{equation}
where $\widetilde{{\bf X}}_{m+1}$ is the forecasted mesh-based velocity field at the next GNN time step (also referred to as a single-step prediction), and ${\bf X}_{m+1}$ is the corresponding target. 

Given successive training snapshot pairs $({\bf X}_m, {\bf X}_{m+1})$ extracted from simulation trajectories, backpropagation-based supervised GNN training can be accomplished by casting objective functions either in terms of target rates ${\bf X}_{m+1}-{\bf X}_m$ or the flowfield ${\bf X}_{m+1}$ directly. The latter approach is leveraged here, producing the mean-squared error (MSE) based objective function 
\begin{equation}
    \label{eq:mse}
    {\cal L}_\text{MSE} = \left \langle \frac{1}{|V| \times N_F} \sum_{i=1}^{|V|} \sum_{j=1}^{N_F} ({\bf X}^{(i,j)}_{m+1} - \widetilde{{\bf X}}^{(i,j)}_{m+1} )^2 \right \rangle. 
\end{equation}
Equation~\ref{eq:mse} represents the single step squared-error in the GNN prediction averaged over the $|V|$ nodes and the $N_F$ target features stored on each node. The angled brackets $\langle . \rangle$ represent an average over a batch of snapshots (i.e., a training mini-batch). The goal of the baseline training procedure is to optimize GNN parameters $\theta$ such that the MSE loss function in Eq.~\ref{eq:mse}, aggregated over all training points, is minimized. Additional details on the training procedure is provided in Appendix~\ref{app:d}. 

Given this optimization goal, it should be noted that the modeling task in Eq.~\ref{eq:gnn_prediction} is challenging by design and deviates from previous mesh-based modeling efforts for a number of reasons:
\begin{enumerate}
    \item \underline{\textbf{Large GNN timestep:}} The flow snapshots are sampled at a fixed GNN timestep $\Delta t$, which is configured here to be significantly larger than the CFD timestep $\delta t$ used to generate the data. When generating the data, the CFD timestep $\delta t$ is chosen to satisfy a stability criterion determined from the numerics used in the PDE solution procedure. Since the ratio $\Delta t / \delta t$ is high (the ratio is 100 here), the GNN is forced to learn a timescale-eliminating surrogate that is much more nonlinear and complex than the $\Delta t = \delta t$ counterpart, particularly in the high Reynolds numbers regime. The tradeoff is that fewer GNN evaluations are required to generate predictions at a target physical horizon time. 
    \item \underline{\textbf{Absence of pressure:}} The Navier-Stokes equations, which are the governing equations for fluid flow, identify the role of local pressure gradients on the evolution of fluid velocity. The GNN in Eq.~\ref{eq:gnn_prediction}, however, operates only on velocity field data, and does not explicitly include pressure or its gradients in the evaluation of dynamics. Instead, the model is tasked to implicitly recover the effect of pressure on the flow through data observations. This is consistent with real-world scenarios in which time-resolved velocity data is accessible (e.g., through optical flow \cite{optical_flow} or laser-based imaging \cite{adrian_piv}), but pressure field data (particularly pressure gradient data) is inaccessible or sparse. {In the context of incompressible flows (the application considered here), it is possible that explicitly including some pressure contribution, such as a pressure gradient as input, could help with predictions. However, further investigation to this end is required, and such modifications may require either calling a conventional pressure solve or developing some kind of surrogate for the pressure solve (as done in Ref.~\cite{tompson2017accelerating}, for example) during inference, essentially making the method physics-informed and perhaps even tied down to an incompressible flow assumption. The intent of this work is to develop a purely data-based approach to mitigate any such assumptions in the model form or objective function.}
    
    \item \underline{\textbf{Mesh irregularity:}} The mesh from which the graph is derived (provided in Fig.~\ref{fig:bfs_mesh} in Appendix~\ref{app:a}) contains regions of high cell skewness and irregularity, particularly in the far-field regions. Since regression data are sampled directly from cell centers, node neighborhoods near irregular mesh regions are characterized by a high variation in edge length scales. Such irregularities are included here to emphasize the model's inherent compatibility with unstructured grids.
\end{enumerate}

\section{Results}
\label{sec:results}

\subsection{{Enhancing} GNNs for Interpretability}
\label{sec:results:fine_tuning}
It is assumed that a black-box graph neural network optimized for the above described forecasting task -- termed the \textit{baseline} GNN, ${\cal G}_B$ -- is available. With ${\cal G}_B$ as the starting point, the {interpretability enhancement} procedure (to be described and demonstrated in this section) produces a new GNN -- termed the \textit{{interpretable}} GNN, ${{\cal G}_I}$ -- that is both physically interpretable and optimized for the same modeling task {as the baseline ${\cal G}_B$}. These interpretability properties are embedded into the baseline through architectural modifications; more specifically, ${{\cal G}_I}$ is created by appending ${\cal G}_B$ with an interpretable {sub-sampling module} that does not modify the baseline parameters. The module provides interpretability through a learnable pooling operation that adaptively isolates a subset of nodes in the context of the mesh-based forecasting task (the objective function in Eq.~\ref{eq:mse}). It will be shown how these nodes can be readily visualized to extract coherent physical features most relevant to this forecasting goal, all while retaining the architectural structure and predictive properties of the baseline model. This approach is illustrated in Fig.~\ref{fig:arch}. 

\subsubsection{Baseline GNN}
The baseline GNN architecture (Fig.~\ref{fig:arch}, top) follows the encode-process-decode strategy \cite{meshgraphnets}, which is a leading configuration for data-driven mesh-based modeling. Here, in a first step, input node and edge features -- which correspond to velocity components and relative physical space distance vectors, respectively -- are encoded using multi-layer perceptrons (MLPs) into a hidden channel dimension, where $N_H$ is typically much larger than $N_F$, the input feature space size. The result is a feature-encoded graph characterized by the same adjacency matrix as the input graph. This encoded graph is then passed to the \textit{processor}, which contains a set of $L$ independently parameterized message passing layers that operate in the fixed hidden node and edge dimensionality $N_H$. 

Message passing layers can take various forms and are the backbone of all GNN architectures, since they directly leverage the graph connectivity to learn complex functions conditioned on the arrangement of graph nodes in neighborhoods \cite{gilmer_2017}. The processor in the baseline architecture in Fig.~\ref{fig:arch} is a general description, in that the {interpretability enhancement} procedure is not tied down to the type of MP layer used. Here, all MP layers correspond to multiscale message passing (MMP) layers, which leverage a series of coarse grids in the message passing operation to improve neighborhood aggregation efficiency \cite{multiscale_meshgraphnets,shivam_topk,lino_gnn}. Details on the MMP layer used here can be found in Appendix~\ref{app:c}.

\subsubsection{Interpretable {Sub-sampling Module}} 
The interpretable {sub-sampling module} is shown in Fig.~\ref{fig:arch} (bottom), and contains a separate set of layers designed to enhance interpretability properties of the black-box baseline. The module is implemented through a skip connection between the first and last MP layers in the baseline processor block, such that module parameters are informed of the action of baseline neighborhood aggregation functions during parameter optimization. There are three components to the module: an adaptive graph pooling layer, an additional MMP layer that acts on the pooled graph representation, and a parameter-free un-pooling layer. 

\begin{figure}
    \centering
    \includegraphics[width=0.4\textwidth]{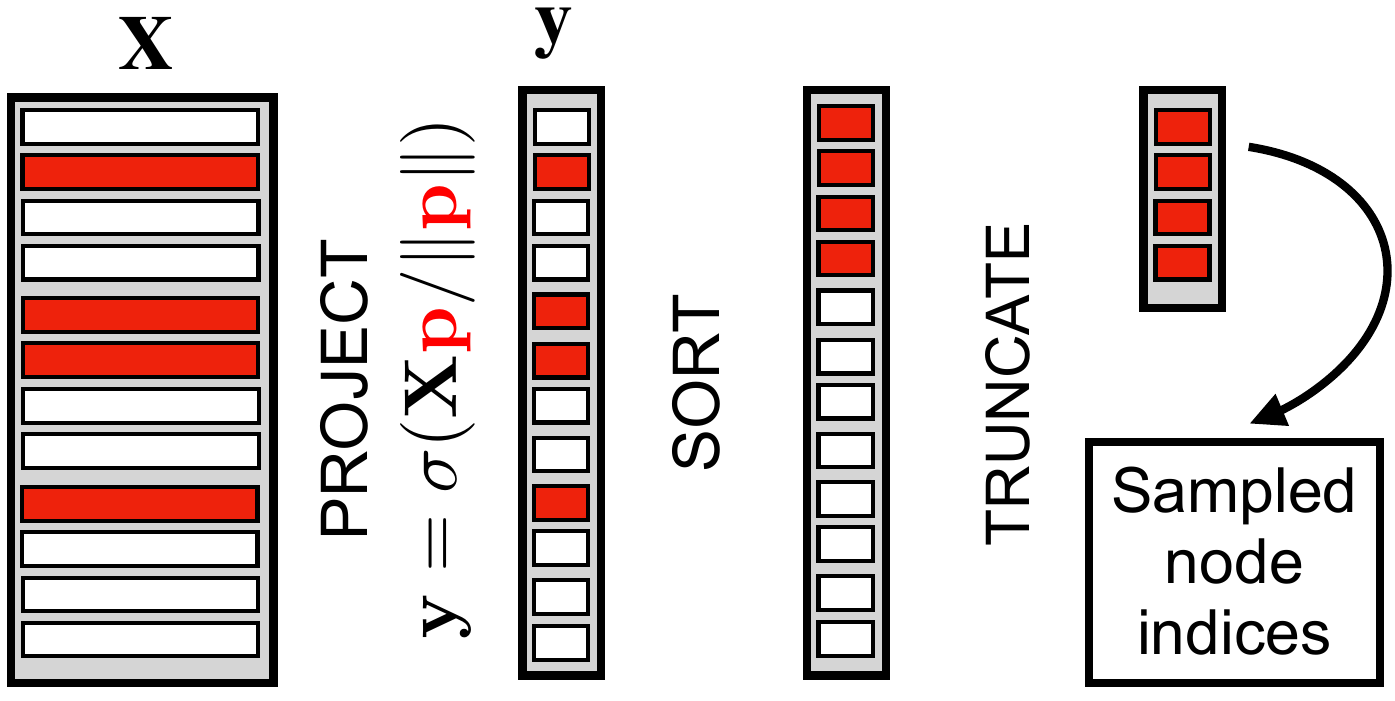}
    \caption{Schematic of Top-K reduction procedure. ${\bf X} \in \mathbb{R}^{|V| \times N_F}$ denotes an input node attribute matrix, and ${\bf y} = \sigma({\bf X}{{\bf p}}/\lVert{{\bf p}}\rVert) \in \mathbb{R}^{|V| \times 1}$ is the result of the node-wise projection operation, where ${\bf p} \in \mathbb{R}^{N_F \times 1}$ is learned during training and $\sigma$ is a hyperbolic tangent activation function.}
    \label{fig:topk}
\end{figure}

Interpretability comes from the graph pooling layer, which uses an adaptive node reduction operation as a mechanism for graph dimensionality reduction. This is accomplished using a Top-K based projection-truncation operation \cite{gao2019graph}, illustrated in Fig.~\ref{fig:topk}. In a first step, the input node attribute matrix is reduced feature-wise to a one-dimensional feature space using a projection vector that is learned during training, which constitutes the parameter of the Top-K layer. In a second step, the nodes in this feature space are sorted in descending order; nodes corresponding to the first $K$ values in the reduced feature space are retained, and the remaining nodes are discarded. The node reduction factor, given by the integer $RF=|V|/K$, is a hyperparameter indicating the degree of dimensionality reduction achieved by the Top-K operation: a higher $RF$ means fewer nodes are retained. The truncation procedure produces a sub-graph coincident with a subset of nodes and edges in the original graph. That is, the retained nodes are a subset of $V$, and the edges belonging to these nodes are a subset of $E$. The MMP layer in the module acts on the neighborhoods in this sub-graph, the output of which is un-pooled and sent back to the baseline processor using a residual connection. 

As illustrated in Fig.~\ref{fig:arch}, the sub-graph produced by the pooling layer can be readily visualized in a mesh-based setting, since the extracted nodes correspond to locations in physical space. If the predictive capability of the baseline is withheld when optimizing the parameters in the interpretable module, nodes identified in the sub-graphs serve as an accessible link to the forecasting task. The discussion below shows how adaptive node sub-sampling provided by the pooling operation identifies physically coherent artifacts in the mesh-based domain, thereby connecting the data-based modeling task (the goal of the baseline) with application-oriented features of interest (the goal of the interpretable {sub-sampling module}). 

\subsubsection{{Interpretability Enhancement} Procedure}
\label{sec:results:fine_tuning:procedure} 
{The interpretability enhancement procedure} is accomplished by first initializing the ${{\cal G}_I}$ model parameters directly from those in the baseline ${\cal G}_B$, which is assumed to already have been trained. Then, the interpretable module is appended to the ${{\cal G}_I}$ architecture as per Fig.~\ref{fig:arch}. 

\begin{figure}
    \centering
    \includegraphics[width=0.5\columnwidth]{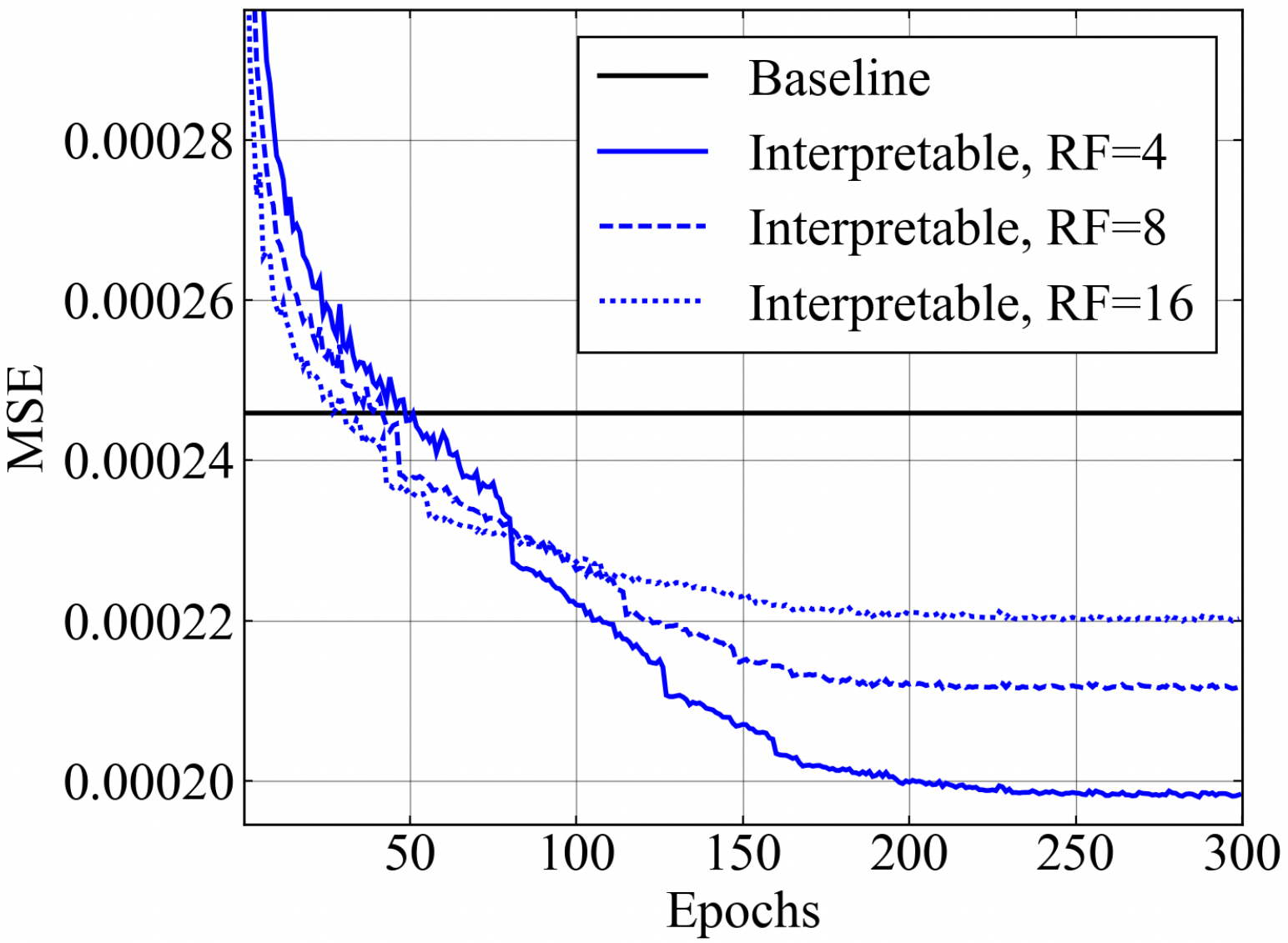}
    \caption{Mean-squared error objective (Eq.~\ref{eq:mse}) versus training epochs during {interpretability enhancement} phase.}
    \label{fig:finetune_training}
\end{figure}

A training step for ${{\cal G}_I}$ is then carried out using the same objective function used to train the baseline ${\cal G}_B$. During training, the subset of parameters in ${{\cal G}_I}$ corresponding to the baseline is frozen, and the parameters corresponding to the interpretable module are freely trainable. {This ensures that the {interpretable GNN} parameters include the baseline parameters as a subset, isolating optimization focus to the interpretable module parameters during training.} The {interpretability enhancement} procedure is shown in Fig.~\ref{fig:finetune_training}, which displays objective function optimization histories for {interpretable GNN}s trained with different values of node reduction factor $RF$. 

The training histories show two key trends: (1) the baseline objective history is frozen, as the baseline model ${\cal G}_B$ is not being modified, and (2) the {interpretable} models achieve lower converged mean-squared errors in single-step GNN predictions than the baseline. The latter trend is consistent with the fact that additional parameters are introduced in the {interpretability enhancement} procedure (see Fig.~\ref{fig:arch}). Due to the fact that a message passing layer acts on the Top-K-identified subgraphs, the converged training errors increase with decreasing RF. Higher RF translates to sub-graphs occupying a smaller percentage of original graph nodes, which in turn limits the scope of the additional message passing layer. 

\subsubsection{Masked Fields} 
\label{sec:results:masked_fields}

\begin{figure}[t!]
    \centering
    \includegraphics[width=0.6\columnwidth]{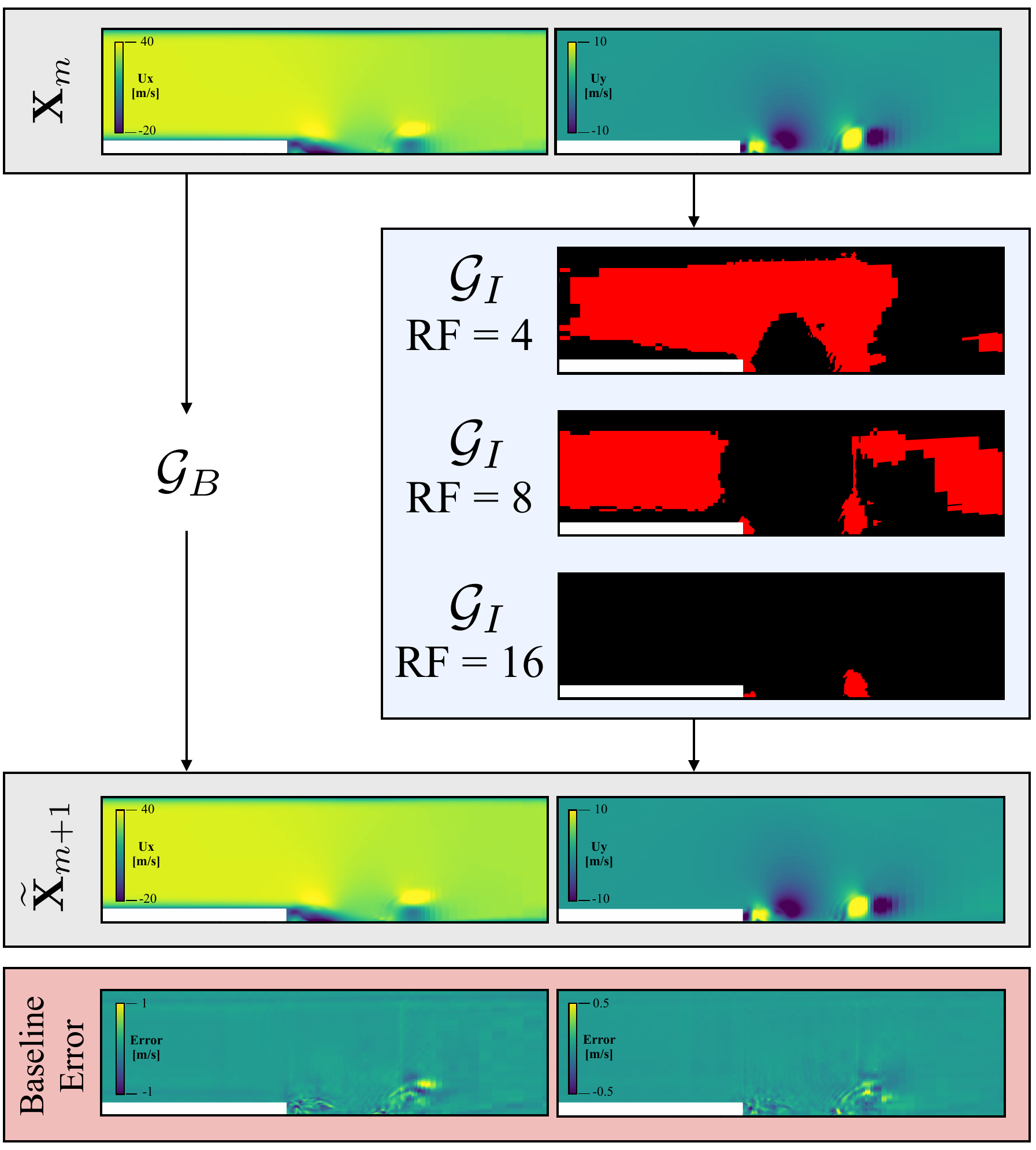}
    \caption{Comparison of GNN prediction workflows between baseline and {interpretable GNN}s. {interpretable} workflow includes masked fields generated in forward pass. Input flowfield (top) sourced from Re=27,233 trajectory of the testing set. Bottom-most plots show baseline absolute error fields ($\widetilde{\bf X}_{m+1} - {\bf X}_{m+1}$) for streamwise (left) and vertical (right) velocity components.}
    \label{fig:example_prediction_finetune}
\end{figure}

The interpretability enhancement is illustrated in Fig.~\ref{fig:example_prediction_finetune}, which contrasts the single-step prediction workflow of the baseline GNN with that of a series of {interpretable GNN}s, each having incrementally higher values of RF. The input to all models is the same instantaneous mesh-based flowfield ${\bf X}_m$ sourced from a testing set trajectory.

Figure~\ref{fig:example_prediction_finetune} shows how the {interpretability enhancement} procedure adds \textit{masked fields} to the model evaluation output. Simply put, a masked field is generated by visualizing the sub-graph node indices produced by the Top-K pooling operation. Since these truncated nodes coincide with physical space locations consistent with the underlying mesh, the ``active" nodes selected during pooling can be extracted and visualized as a node-based categorical quantity (1 if the node was identified and 0 otherwise). 

The features contained in these fields allow one to interpret the relationship between the input flow field ${\bf X}_m$ and the model-derived forecast $\widetilde{\bf X}_{m+1}$. In other words, through visualization of the masked fields, the {interpretable} models allow one to access regions in physical space intrinsically linked to the modeling task (single-step prediction), thereby making the model behavior directly accessible from the perspective of the objective function used during training. The masked fields in Fig.~\ref{fig:example_prediction_finetune} show how this added capability translates to interpretable prediction unavailable in the baseline, as the masked fields interestingly isolate nontrivial, but spatially coherent, clusters of nodes (often disjoint in physical space) within the GNN forward pass. Such features can then be connected to the problem physics in an expert-guided analysis phase. 

For example, as the reduction factor increases, the masked fields in Fig.~\ref{fig:example_prediction_finetune} isolate coherent regions corresponding to near-step recirculation zones and downstream vortex shedding. In particular, the increase in $RF$ emphasizes importance of the step cavity region in the forecasting task -- the emphasis on the free-stream region in the masked field drops, implying that the role played in the step cavity region is much more important when making accurate model forecasts. This trend is expected in the BFS configuration, and is overall consistent with the fact that high baseline model errors are also observed in this region. 
\begin{figure}
    \centering
    \includegraphics[width=\textwidth]{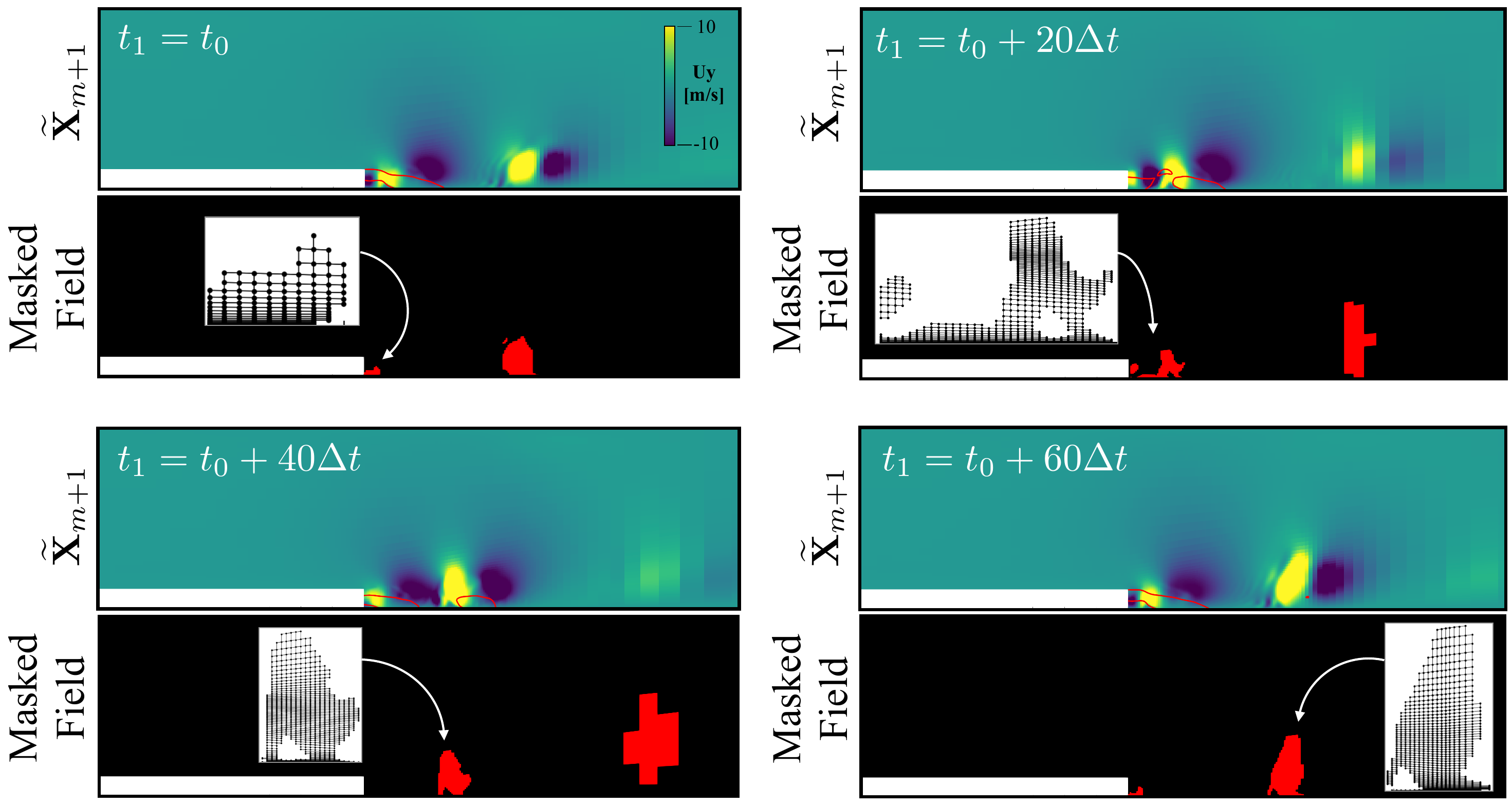}
    \caption{Single-step prediction outputs generated by {interpretable GNN} (RF=16) using input fields at time-ordered instants in the Re=27233 trajectory. Each frame shows predicted y-component of velocity (top) alongside masked field (bottom), with inlays showing subgraph representations.}
    \label{fig:masked_field_evolution}
\end{figure}

Since the input velocities stored on the nodes are time-evolving, and the Top-K projection operation is conditioned on these input values, the regions identified in the masked fields are necessarily time-evolving as well. Consequently, the masked fields identify coherent structures that evolve in accordance with the unsteady features of the flow -- in the BFS configuration, these features correspond to emerging and downstream shedding of recirculation zones, which in turn lead to a shedding cycle that affects the location at which the flow reattaches to the bottom wall. This time-dependent quality is highlighted in Fig.~\ref{fig:masked_field_evolution}, which shows a set of time-ordered single-step predictions paired with corresponding masked fields during one BFS shedding cycle. It should be noted that the adaptive pooling operation induces a sub-graph from which the masked field is derived; since the identified regions in the masked fields are time-evolving, the adjacency matrix that characterizes this sub-graph is time-evolving as well, and remains coincident with the fixed adjacency matrix of the original mesh-based graph. The sub-graphs are provided as inlaid plots in Fig.~\ref{fig:masked_field_evolution}; these evolving adjacency matrices enable successive utilization of MMP layers on these isolated regions in the {interpretable} model.

\subsubsection{Preserving Baseline Performance} Incorporation of the interpretability enhancement does not come for free, and incurs the following costs: (1) there is an additional offline training step; and (2) inference of the {interpretable} models is more costly compared to their baseline counterparts, due to the introduction of an additional set of arithmetic operations through the interpretable module attachment. In light of these costs, the advantage is that the {interpretable} models also maintain the performance of the baseline models in terms of forecasting accuracy and stability characteristics. 

\begin{figure}
    \centering
    \includegraphics[width=0.5\columnwidth]{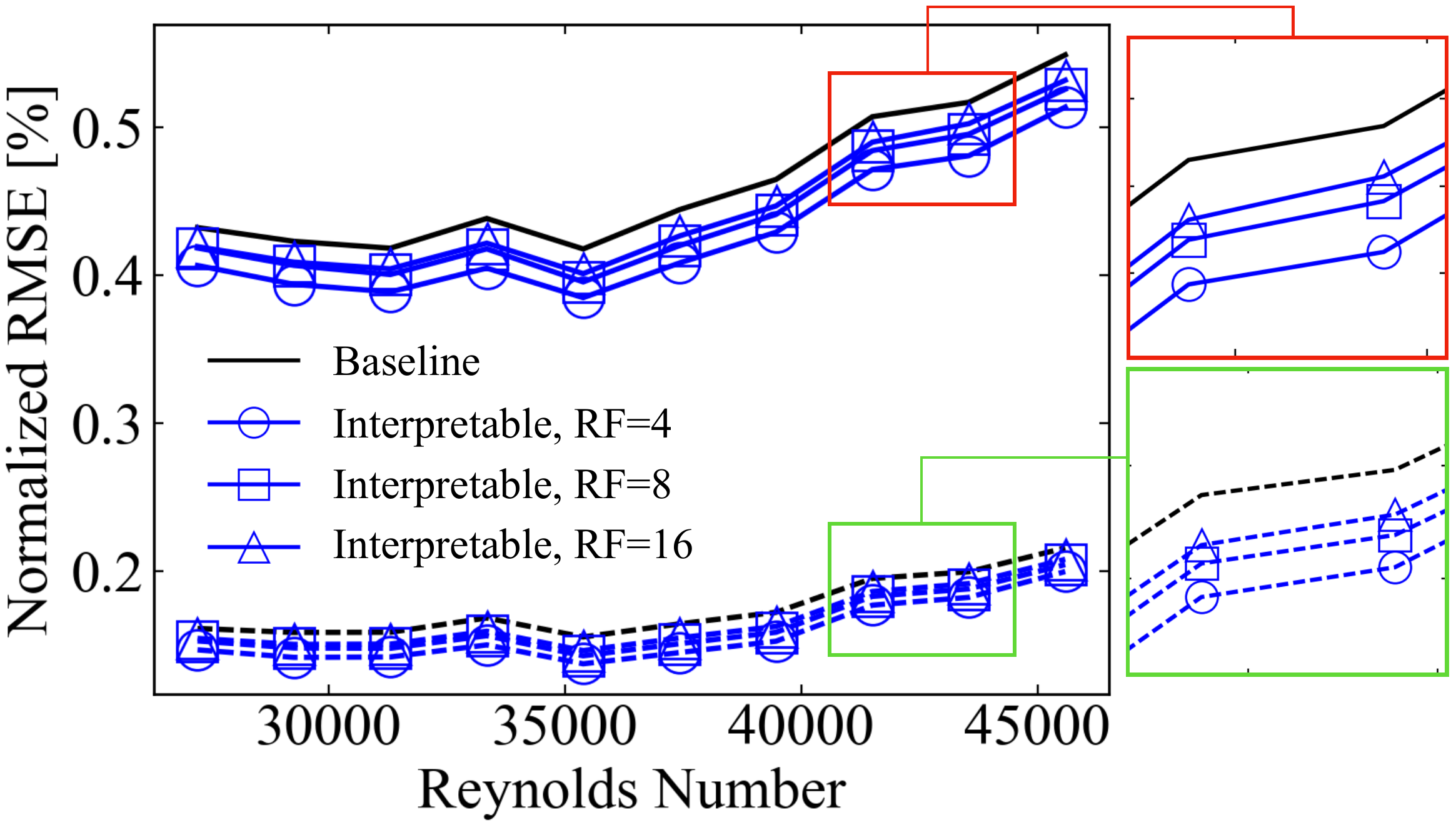}
    \caption{Normalized RMSE (Eq.~\ref{eq:rmse}) versus testing Reynolds number for streamwise (solid lines) and vertical (dashed lines) velocity components.}
    \label{fig:single_step_error}
\end{figure}

To emphasize this point, Fig.~\ref{fig:single_step_error} displays root-mean-squared errors (RMSE) in single-step predictions for baseline and {interpretable} models as a function of testing set Reynolds numbers. The RMSE is calculated node-wise, is computed separately for each predicted output feature, and is normalized by the inlet velocity to allow for relative comparisons across the range of tested Reynolds numbers. More formally, for the $j$-th output feature (the velocity component), it is defined as the relative error percentage
\begin{equation}
    \label{eq:rmse}
    \text{RMSE}_j =  \frac{\left \langle \sqrt{\frac{1}{|V|} \sum_{i=1}^{|V|} ({\bf X}^{(i,j)}_{m+1} - \widetilde{{\bf X}}^{(i,j)}_{m+1} )^2 } \right \rangle}{u_{in}} \times 100 , 
\end{equation}
where $\widetilde{{\bf X}}^{(i,j)}_{m+1}$ denotes the single-step GNN prediction for the $j$-th feature at graph node $i$, ${{\bf X}}^{(i,j)}_{m+1}$ is the corresponding target, $u_{in}$ is the inlet velocity corresponding to a particular Reynolds number, and the angled brackets $\langle . \rangle$ denote an ensemble average over testing set samples. 

Since the {interpretability enhancement} procedure discussed thus far leverages the same objective function as the baseline -- namely, the single-step forecasting error in Eq.~\ref{eq:mse} -- the trends in testing set evaluations for this metric are overall consistent with what was observed during training (see Fig.~\ref{fig:finetune_training}). More specifically, Fig.~\ref{fig:single_step_error} shows how the baseline is effectively an upper bound for the model error at all evaluation Reynolds numbers. Additionally, the error in {interpretable} models approaches the baseline as the node reduction factor $RF$ increases. It should also be noted that the normalized error curves are not flat -- after about Re=35,000, the relative errors begin to increase, emphasizing the increase in unsteady flow complexity in this regime. This increase is likely an artifact of the fixed timestep using during training. Alongside higher velocity magnitudes fluctuation in the near-step regions, since the shedding cycle frequencies are more rapid at higher Reynolds numbers, the forecasting task becomes more difficult in a given fixed GNN $\Delta t$ window.

\begin{figure}
    \centering
    \includegraphics[width=\textwidth]{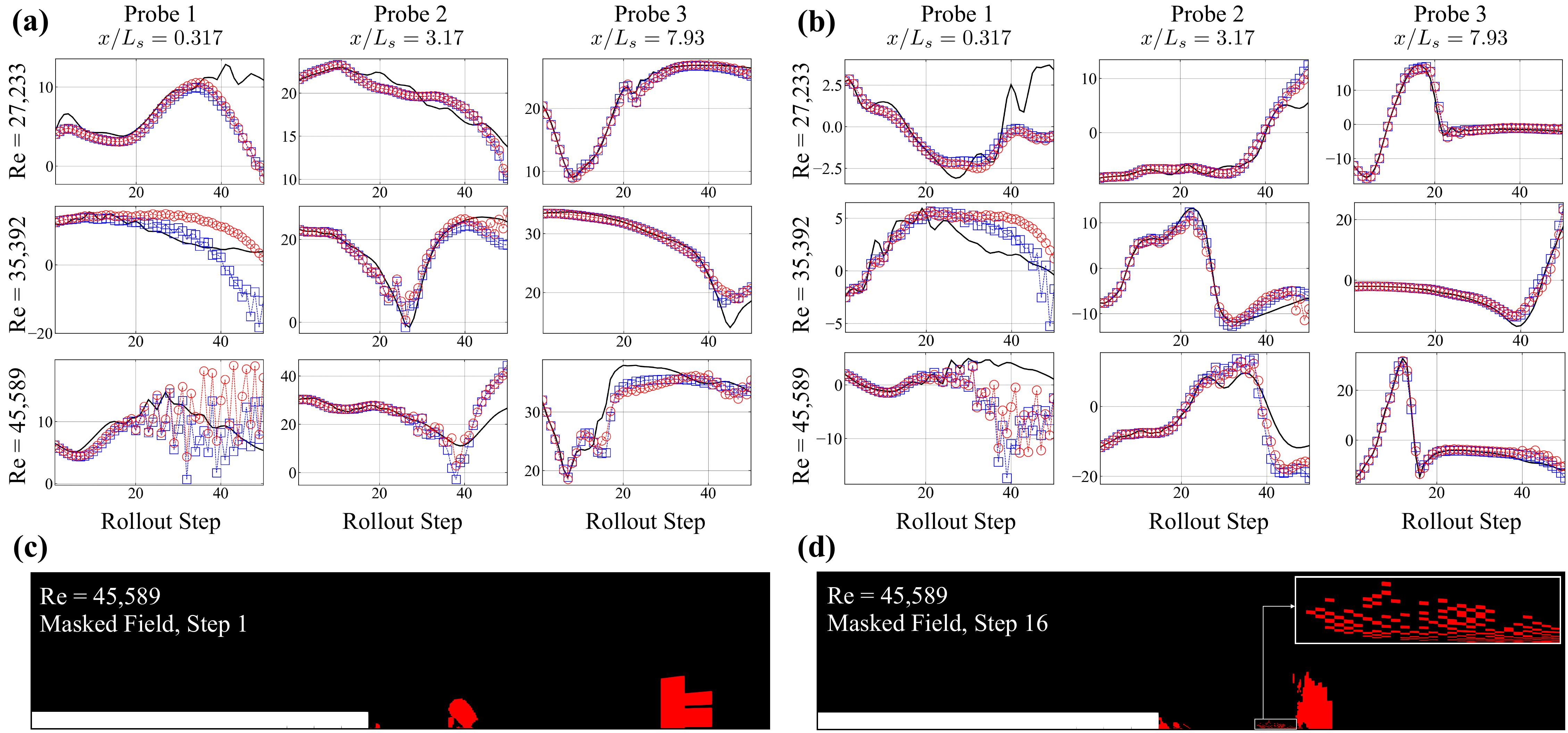}
    \caption{\textbf{(a)} Rollout predictions for streamwise velocity component at fixed probe locations for baseline GNN (blue squares) and {interpretable GNN} (RF=16, red circles). Solid black line indicates target value. Rows correspond to different initial conditions sourced from the indicated Reynolds numbers; columns correspond to different probe locations at specified x-locations away from step anchor (y-locations of probes coincide with the step height). \textbf{(b)} Same as (a), but for vertical velocity component. \textbf{(c)} Masked field after 1 rollout step for Re=45,589. \textbf{(d)} Masked field after 16 rollout steps for Re=45,589; inlay shows zoom-in on a portion of the masked field.}
    \label{fig:rollout_plots}
\end{figure}

Although the averaged single-step error is valuable when validating the consistency between baseline and {interpretable} model outputs in an a-priori setting, comparison of model performance in an a-posteriori (or rollout\footnote{Here, the model is evaluated in an autoregressive context, and therefore accounts for the accumulation of GNN error throughout a series of prediction timesteps (or rollout steps).}) setting is required to confirm that the {interpretability enhancement} procedure indeed preserves the overall characteristics of the baseline model in a truly predictive context. Such a comparison is given in Fig.~\ref{fig:rollout_plots}(a) and (b), which shows rollout predictions for streamwise and vertical velocity components respectively for three different Reynolds numbers. The rollouts are sampled at three fixed spatial (probe) locations in the step cavity region, where the BFS dynamics are most complex: probe 1 is the closest to the step anchor, and reflects the impact of flow separation on the unsteady dynamics, whereas probes 2 and 3 are successively further downstream. 

The rollouts ultimately confirm predictive consistency between the {interpretable GNN} and the corresponding baseline -- in almost all cases, the predictions provided by the {interpretable} model very closely resemble those of the baseline. Interestingly, the rollouts in Fig.~\ref{fig:rollout_plots} reveal noticeable error buildup in the near-step regions as opposed to further downstream. The degree of error buildup is most prominent at the highest tested Reynolds number, where the onset of instability starts at about 20 rollout steps; this is consistent with the trends in Fig.~\ref{fig:single_step_error}, which implies that accumulation of single-step error is most pronounced at the highest Reynolds numbers. 

While the rollout accuracy, particularly near the step, is less than ideal, two important points must be stressed: (a) the GNN timestep $\Delta t$ is prescribed here to be significantly larger than the CFD timestep (as discussed in the previous section); and (b) error accumulation is expected since the objective function used during training and {interpretability enhancement} only accounts for the single-step error. Although inclusion of multiple rollout steps during training has been shown in recent work to significantly improve GNN prediction horizons for baseline models \cite{graphcast}, such investigations into long-term stability improvements were not explored in this work, as they significantly increase training times and require complex application-specific training schedules. Since the scope of this work is {interpretability enhancement} with respect to a baseline, the objective of Fig.~\ref{fig:rollout_plots} is to highlight how appending the interpretable {sub-sampling module} to the baseline GNN does not incur significant accuracy penalties in both single-step and rollout contexts. As a result, it can be concluded that the added benefit of masked field interpretability is a meaningful addition to the black-box model.

Figure~\ref{fig:rollout_plots}(c) and (d) provide masked field visualizations from the Re=$45,589$ predictions produced by the {interpretable GNN}s, and show the effect of rollout error accumulation on the masked field structures. More specifically, the masked field at the first rollout step in Fig.~\ref{fig:rollout_plots}(c) isolates largely coherent structures in the step cavity, as observed in the previous examples (see Fig.~\ref{fig:masked_field_evolution}), whereas the masked field after 16 rollout steps (Fig.~\ref{fig:rollout_plots}(d)) includes the emergence of high frequency checkerboard-like patterns that are largely non-physical. These visualizations show a direct correspondence between rollout errors and spatial coherency observed in the identified masked field structures -- although more investigation to this end is required, the advantage is that the masked field is generated directly by the model during inference, and it is possible that such high-frequency artifacts can be used to signal oncoming model error accumulation in an a-posteriori context.

\subsection{Error-Identifying GNNs}
\label{sec:results:budget}
The results above showcased the interpretability enhancement provided by the {interpretable} model from the angle of feature extraction. More specifically, the structures identified by the {interpretable GNN}s through the masked fields serve as an accessible link between the model architecture and the optimization goal (the mesh-based flow forecast). 

The goal of this section is to build on the above by adding more capability to the identified masked fields from the modeling perspective (i.e., providing utility beyond feature extraction). This is accomplished by adding a regularization term, representing a mean-squared error budget, to the objective function during the {interpretability enhancement} process. It is shown below how the inclusion of this relatively simple regularization \emph{forces the identified nodes in the masked field to not only produce coherent features, but also to coincide with regions that contribute most significantly to the {interpretable GNN} forecasting error}, a novel capability not guaranteed in the standard {interpretability enhancement} procedure.

This strategy is inspired from error tagging methods leveraged in adaptive mesh refinement (AMR) based CFD solvers \cite{berger_colella}, in which the first step consists of tagging subset of mesh control volumes expected to contribute most to the PDE discretization error. In AMR strategies, knowledge of error sources in discretization methods used to solve the Navier-Stokes equations is used to guide which regions are tagged (for example, regions containing high gradients in pressure and fluid density); since the baseline GNN forecasting model is considered a black-box, the goal here is to offer an analogous data-based pathway that isolates sub-graphs in input graphs processed by GNNs coinciding with regions of high model error. Such information adds error-tagging utility to the masked fields, which can then be used to guide downstream error-reduction strategies. 

\subsubsection{Error Identification via Regularization}
A novel error identification strategy is accomplished using a budget regularization that modifies the objective function used during the {interpretability enhancement} procedure. This modified objective function adds a regularization term to the standard single-step prediction objective of Eq.~\ref{eq:mse}, and is given by 
\begin{equation}
    \label{eq:reg_loss}
    {\cal L}_{R} = {\cal L}_{\text{MSE}} + \lambda {\cal L}_{B}, 
\end{equation}
where ${\cal L}_{R}$ denotes the regularized loss, ${\cal L}_{\text{MSE}}$ denotes the mean-squared error based forecasting objective in Eq.~\ref{eq:mse}, $\lambda$ is a scaling parameter, and ${\cal L}_B$ is the so-called budget regularization term given by
\begin{equation}
    \label{eq:budget}
    {\cal L}_{B} = \frac{1}{\langle {\cal B} \rangle}, 
    \quad {\cal B} = 
    \frac{\sum_{i=1}^{|V|} \sum_{j=1}^{N_F} {\bf M}_m^i({\bf X}^{(i,j)}_{m+1} - \widetilde{{\bf X}}^{(i,j)}_{m+1} )^2}
    {\sum_{i=1}^{|V|} \sum_{j=1}^{N_F} ({\bf X}^{(i,j)}_{m+1} - \widetilde{{\bf X}}^{(i,j)}_{m+1} )^2}.
\end{equation}

\begin{figure}
    \centering
    \includegraphics[width=\textwidth]{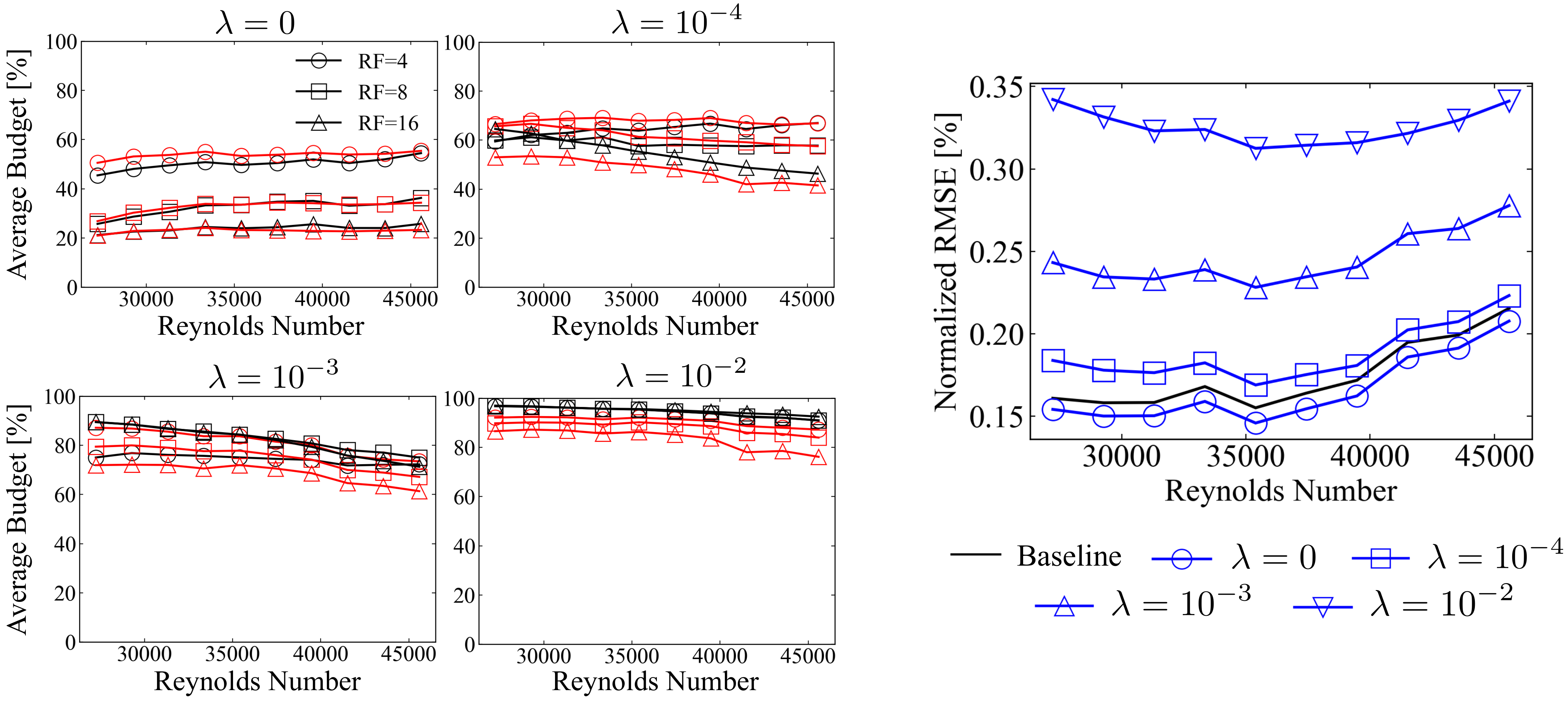}
    \caption{\textbf{(Left)} Average masked field budgets (Eq.~\ref{eq:budget}) versus testing Reynolds number produced by {interpretable GNN}s using increasing values of $\lambda$. Black curves correspond to streamwise velocity component, and red to vertical velocity component. \textbf{(Right)} Normalized RMSE (Eq.~\ref{eq:rmse}) versus testing Reynolds number for baseline GNN (black) and {interpretable GNN}s trained with different levels of budget regularization scaling factor $\lambda$ (markers). All {interpretable GNN}s in figure use $RF=16$. RMSE curves shown for vertical velocity component (trends are consistent for streamwise component).}
    \label{fig:average_budget_rmse_lambda}
\end{figure}

In Eq.~\ref{eq:budget}, the quantity ${\cal B}$ is the masked field budget, and represents the contribution to the total mean-squared error for the single-step forecast from the nodes identified in the masked field ${\bf M}_m^i$, where the subscript $m$ denotes dependence on the input flowfield at the $m$-th time step (${\bf X}_m$), and the superscript $i$ indexes a node in the graph. As per the definition of the masked field, ${\bf M}_m^i$ is 1 if the node is identified through the Top-K pooling procedure, and 0 otherwise. Since the regularization term ${\cal L}_B$ is cast as the inverse of the budget ${\cal B}$, a model with a small value of ${\cal L}_B$ indicates that a large amount of forecasting error is contained in the nodes identified through the masked field. The parameter $\lambda$ in Eq.~\ref{eq:reg_loss} dictates the influence of the regularization term during the {interpretability enhancement} stage, and therefore plays a critical role. Note that setting $\lambda = 0$ recovers the non-regularized {interpretability enhancement} strategy discussed throughout Sec.~\ref{sec:results:fine_tuning}. Through the dependence of ${\cal B}$ on the masked field, the objective is to ensure that -- if the regularization strategy succeeds in reducing ${\cal L}_B$ -- the nodes responsible for a majority of the forecasting error are accessible as an explicit function of the input mesh-based flowfield ${\bf X}_m$, resulting in a type of a-posteriori error tagging strategy. 

\subsubsection{Impact of Regularization}

The effect of the scale factor $\lambda$ on the budget-regularized {interpretability enhancement} is shown in Fig.~\ref{fig:average_budget_rmse_lambda} (left), which plots average budget values (single-step prediction error contained in the masked fields) conditioned on both testing set Reynolds number and velocity component for various {interpretable GNN} configurations. The figure illustrates (a) how various degrees of error tagging capability are indeed being provided through the budget regularization procedure in direct proportion to $\lambda$, and (b) the expected relationship between the node reduction factor and budget (higher $RF$ gives lower budget contributions). For example, the standard {interpretability enhancement} strategy corresponding to $\lambda = 0$ produces masked fields that describe a noticeably lower amount of model error than the regularized ($\lambda > 0$) counterparts. Interestingly, the implication is that without budget regularization, the generated masked fields -- even for a relatively large node reduction factor of $RF=4$, meaning that $25\%$ of the total nodes are sampled -- account for only roughly half of the forecasting error contribution. On the other hand, as $\lambda$ is progressively increased from $10^{-4}$ to $10^{-2}$, the budget values increase across the board. The $\lambda = 10^{-2}$ case nearly reaches a saturation point, where a majority of the single-step forecasting error in most cases is described by the identified nodes. The budget curves are almost flat, with some exceptions at the higher tested Reynolds numbers, implying that the regularization strategy learns a robust and interpretable error-identifying mechanism.

Although the increase in $\lambda$ translates to an increase in error-identifying capability in the {interpretable GNN}s, Fig.~\ref{fig:average_budget_rmse_lambda} (right) shows how the regularization incurs an accuracy tradeoff. More specifically, Fig.~\ref{fig:average_budget_rmse_lambda} (right) shows effects of the $\lambda$ parameter on the {interpretable GNN}s in terms of single-step normalized RMSE values (Eq.~\ref{eq:rmse}) conditioned on the same testing set Reynolds numbers. When the budget regularization is introduced, the scaling parameter effectively increases the forecasting errors for all observed Reynolds numbers, implying a type of accuracy tradeoff with respect to the additional modeling capability provided by the budget regularization. This is overall consistent with the $\lambda$-scaling process, in that a greater emphasis on minimizing the inverse budget via high $\lambda$ necessitates less optimization focus on the direct forecasting error objective. This attribute is directly apparent in Fig.~\ref{fig:average_budget_rmse_lambda} (right) -- as $\lambda \rightarrow 0$, the RMSE curves converge to the standard {interpretable GNN}, which achieves lower forecasting error relative to the baseline GNN used to initiate the {interpretability enhancement} procedure.

\begin{figure}
    \centering
    \includegraphics[width=0.6\columnwidth]{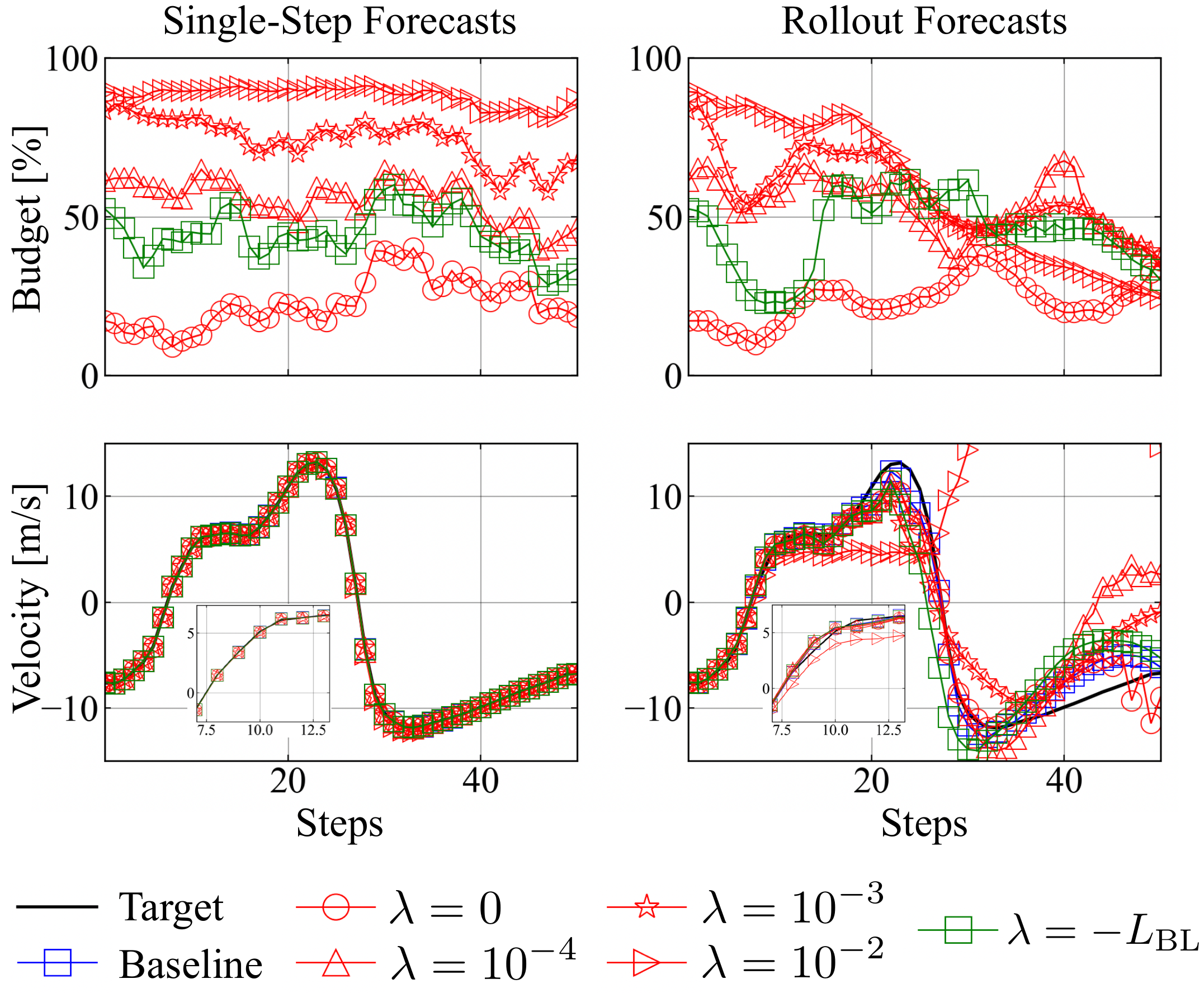}
    \caption{\textbf{(Top row)} Masked field budgets (Eq.~\ref{eq:budget}) versus time step for single-step (left plot) and rollout (right plot) predictions using initial condition from Re=35,392 testing set trajectory. {interpretable GNN}s have RF=16. \textbf{(Bottom row)} Corresponding velocity field predictions (vertical component) sampled at a single spatial location (probe 2, same as used in middle row of Fig.~\ref{fig:rollout_plots}(b)). The $\lambda=-L_{\text{BL}}$ setting (green squares) is discussed in Sec.~\ref{sec:results:optimal_lambda}.} 
    \label{fig:budget_rollout}
\end{figure}

To further assess the effect of budget regularization in a more practical inference-based setting, Fig.~\ref{fig:budget_rollout} displays forecasting results in terms of masked field budget and predicted velocity versus time step using an initial condition sourced from the Re=35,392 testing set trajectory. The figure compares {interpretable GNN} forecasts in single-step and rollout prediction configurations, with models corresponding to node reduction factors (RFs) of 16 (the adaptive pooling procedure isolates sub-graphs comprised of 16x fewer nodes than the original). 

The single-step forecasts largely mimic the trends observed in Fig.~\ref{fig:average_budget_rmse_lambda} (left), in that the regularization imposed during {interpretability enhancement} manifests through an increase in error budget during inference with higher $\lambda$. These single-step forecasts, which assume no accumulation of model forecasting error through forward time steps, highlight the control provided by the regularization procedure by the $\lambda$ parameter. More specifically, Fig.~\ref{fig:budget_rollout} indicates diminishing returns in the regularization procedure with respect to $\lambda$: for example, the jump in budget when moving from $\lambda=0$ to $\lambda=10^{-4}$ is on average higher than when moving from $\lambda=10^{-3}$ to $\lambda=10^{-2}$. The rollout trends, which do account for the accumulation of single-step forecasting error through trajectory generation, further point to this aspect of diminishing returns, and directly reveal how the added emphasis on budget regularization through higher $\lambda$ values incurs an accuracy tradeoff. For example, the {interpretable GNN} trained with $\lambda=10^{-2}$ departs from the target solution much earlier than the others, which is consistent with the fact that it places more emphasis on minimizing the inverse budget during {interpretability enhancement} -- ultimately, the cost of having a more ``useful" masked field through the ability to tag regions of high error is a drop in forecasting accuracy and stability. Interestingly, although the correlation between budget and the $\lambda$ parameter is present for the first few rollout steps, this correlation decays once error starts to pollute the predictions. In other words, the budgets in the regularized models eventually tend towards the budgets produced by the non-regularized model. Since the baseline and {interpretability enhancement} procedures utilize single-step training objectives, the impact of error accumulation after a single forecasting step is high -- as mentioned earlier, these effects can be mitigated with more robust (albiet more expensive) training strategies that account for the error of more than one future time-step in the objective function.

\begin{figure}
    \centering
    \includegraphics[width=\textwidth]{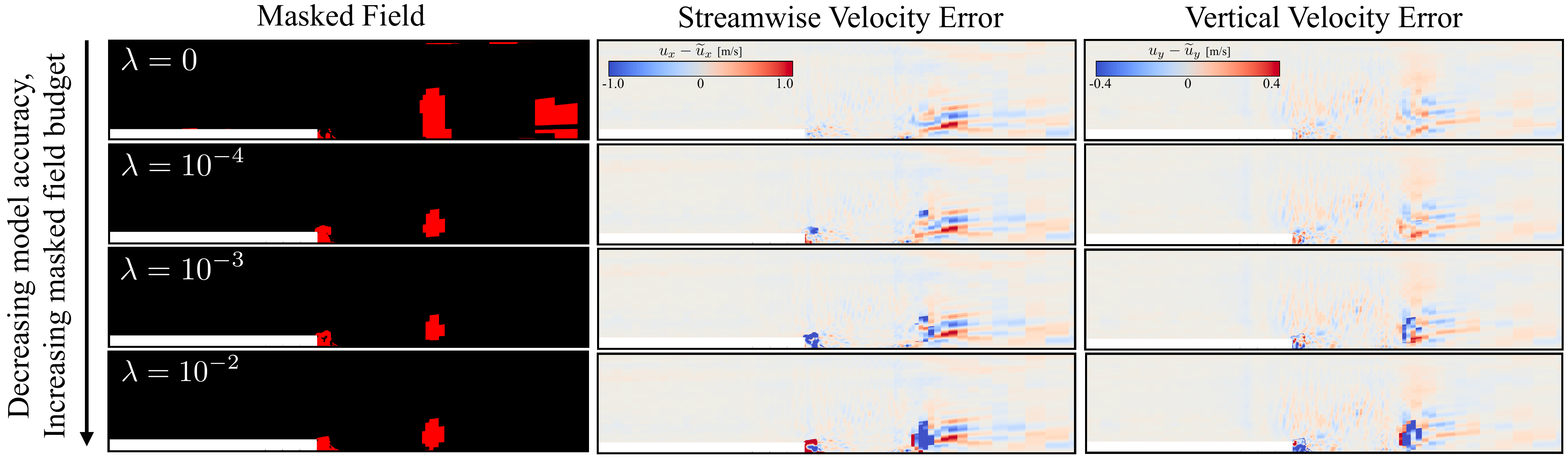}
    \caption{\textbf{(Left)} Masked fields produced by {interpretable GNN}s (RF=16) of increasing $\lambda$ settings for a given input ${\bf X}_m$ (input comes from trajectory described in Fig.~\ref{fig:budget_rollout}). \textbf{(Middle)} Streamwise component of the forecast error (${\bf X}_{m+1} - \widetilde{\bf X}_{m+1}$) after a single step prediction for each of the corresponding {interpretable} models. \textbf{(Right)} Vertical component of the error.}
    \label{fig:budget_rollout_mask}
\end{figure}

It should be noted that for small values of $\lambda$ (e.g., $\lambda=10^{-4}$ in Fig.~\ref{fig:budget_rollout}), the regularization in Eq.~\ref{eq:reg_loss} produces error-identifying masked fields at relatively minimal expense in terms of model accuracy with respect to the baseline GNN, thereby offering a unique interpretable modeling advantage from the perspective of error tagging. This advantage is more apparent when considering Fig.~\ref{fig:budget_rollout_mask}, which shows masked fields produced by the {interpretable GNN}s used to make the predictions in Fig.~\ref{fig:budget_rollout}. The introduction of $\lambda$ incurs a noticeable change to the identified structures in the masked field; more specifically, the non-regularized $\lambda=0$ {interpretable GNN}s isolate structures in the far downstream region, whereas the regularized models isolate more nodes in the near-step vicinity, particularly near the corner at which flow separation is initiated. The difference in masked field structures emphasizes the difference in {interpretable GNN objectives}, in that the identified nodes in the $\lambda>0$ models represent regions in which the predicted mesh-based flowfield $\widetilde{X}_{m+1}$ will contain a majority of the model error. This is evidenced in the corresponding error fields, which showcases the effect of budget regularization on the localization of forecasting error towards the masked regions. 

Additionally, the structures in the masked fields from $\lambda=10^{-4}$ to $\lambda=10^{-2}$ in Fig.~\ref{fig:budget_rollout_mask} are largely similar, which allows the user to tune the degree of error identification capability provided by the {interpretable GNN}. For example, the $\lambda=10^{-4}$ model offers (a) comparable performance to the baseline GNN, (b) interpretability through the identification of coherent structures in the masked field, a utility also provided in the non-regularized $\lambda=0$ model, and (c) the ability to attribute the identified regions to expected sources of forecasting error. Although this latter capability is limited to the scope of the prediction horizon of the model (this is apparent through the rollout results in Fig.~\ref{fig:budget_rollout}), the inclusion of the budget-based regularization reveals a promising pathway for understanding and isolating sources of model error during inference through the adaptive pooling strategy. 

\subsubsection{Optimal Selection of Regularization Factor}
\label{sec:results:optimal_lambda}

The discussion above reveals the key role played by the regularization factor $\lambda$ as a controller for the trade-off between error tagging capability (quantitatively measured by the mean-squared error budget $\cal B$) and predictive accuracy of the {interpretable GNN}. The presence of such a trade-off -- a direct by-product of the form of the regularized objective function in Eq.~\ref{eq:reg_loss} -- naturally leads to the notion of whether or not an ideal (or optimal) $\lambda$ that balances accuracy with error-tagging capability can be determined both a-priori and in a problem-independent manner. In particular, the following question comes into play: for a given Top-K reduction factor, is it possible to determine a value for $\lambda$ that maximizes the error budget contained in masked field (i.e., maximizes error-tagging capability) \textit{while also retaining the predictive capability of the baseline GNN}? A $\lambda$-selection strategy to this end is described below. 

To guide selection of $\lambda$, the expression in Eq.~\ref{eq:reg_loss} can be simplified under two limiting conditions, both of which represent idealized performance targets of the regularized {interpretable} model. The first condition coincides with maximizing the MSE budget: in an ideal error-tagging scenario, 100\% of the error in the forecast should be contained in the nodes identified by the masked field (a budget $\cal B$ of 1). In the ${\cal B} \rightarrow 1$ limit, the budget regularization loss ${\cal L}_{\cal B}$ also becomes 1. The second condition operates under the assumption of zero error in the {interpretable} model; in other words, an ideal model in the predictive sense would also recover ${\cal L}_R = 0$ (see Eq.~\ref{eq:reg_loss}). Under these two limiting conditions, a simplified expression for the modeling objective in Eq.~\ref{eq:reg_loss} can be obtained as  $0 = {\cal L}_{\text{MSE}} + \lambda$, which in-turn provides an estimation for an "ideal" $\lambda$ as $\lambda = -L_{\text{MSE}}$. Since a baseline model is available, the value $L_{\text{MSE}}$ can be set to $L_{\text{BL}}$, the converged MSE loss of the trained baseline GNN. 

Although a negative value for $\lambda$ ($\lambda = -L_{\text{BL}}$) appears counter-intuitive, the relationship stems directly from the ideal modeling limits described above. Since $\lambda$ is negative, the budget regularization term during {interpretability enhancement} is adjusted to the direct budget instead of the inverse budget (i.e., ${\cal L}_B = {\langle {\cal B} \rangle}$ instead of ${\cal L}_{B} = \frac{1}{\langle {\cal B} \rangle}$) to penalize large negative values for $\lambda {\cal L}_{B}$ during training. Although not shown here, the training loss histories for the MSE contribution (${\cal L}_{\text{MSE}}$, the single-step forecast error) validate this approach, in that the corresponding {interpretable} model in the $\lambda = -L_{\text{BL}}$ setting was found match the baseline prediction error on the training set.

For a more practical assessment, testing set budget evaluations for the $\lambda = -L_{\text{BL}}$ case are overlaid in Fig.~\ref{fig:budget_rollout} (green squares). In the RF=16 case, the figure shows how the new model achieves budgets sustained near 50\%. These budgets are comparable to the lowest tested $\lambda=10^{-4}$ model (this is likely due to the fact that $L_{\text{BL}} = 2.46 \times 10^{-4}$, and is therefore at the same order-of-magnitude) and are consistently higher than the un-regularized $\lambda=0$ model. The benefit of the $\lambda = -L_{\text{BL}}$ setting, however, becomes apparent when assessing the prediction errors (bottom row of Fig.~\ref{fig:budget_rollout}). The fact that the forecasting behavior is qualitatively very similar to the baseline model (even at higher rollout steps) illustrates the key advantage provided by this $\lambda$-selection strategy: if one seeks to \textit{maintain} baseline GNN performance while explicitly adding in error-tagging capability through the budget regularization, the $\lambda=-L_{\text{BL}}$ selection introduced here is a promising pathway. 

\subsection{Masked Fields on Extrapolated Geometries}
\label{sec:results:geom_extrap}
The results in Sec.~\ref{sec:results:fine_tuning} show how the {interpretability enhancement} procedure provides the GNN the ability to generate identifiable structures through the masked fields. The results in Sec.~\ref{sec:results:budget} build on this intuition: through a budget regularization procedure conditioned on the identified masked fields, it was shown how a baseline GNN can be explicitly {enhanced} for interpretable error-tagging. Although these capabilities have many useful modeling implications (alongside the direct feature extraction benefits of the masked fields), trends in the previous sections were assessed only from the perspective of a single geometry, namely the backward-facing step.   

The objective of this section is to expand on the previous analyses by means of \textit{geometry extrapolation}, which serves to not only verify masked fields trends, but also highlights the principal advantage of fully graph-based architectures: their ability to be used on unseen meshes. Here, {interpretable GNN}s trained on the BFS geometry are evaluated on two additional unseen geometries exhibiting different flow separation physics. These new geometries are the angled ramp and mounted cube configurations, which constitute a more challenging extrapolation task than, for example, extrapolating on another BFS geometry with the step positioned in a different part of the domain (the latter does not change the mechanism of flow separation, but rather only its spatial location). Fig.~\ref{fig:ramp_cube_geometry} in Appendix~\ref{app:a} provides images of the meshes used for the new geometries. Within the ramp and cube extrapolation context, the goal is to assess consistency in the masked fields from two angles, described below. 

\begin{figure}
    \centering
    \includegraphics[width=\textwidth]{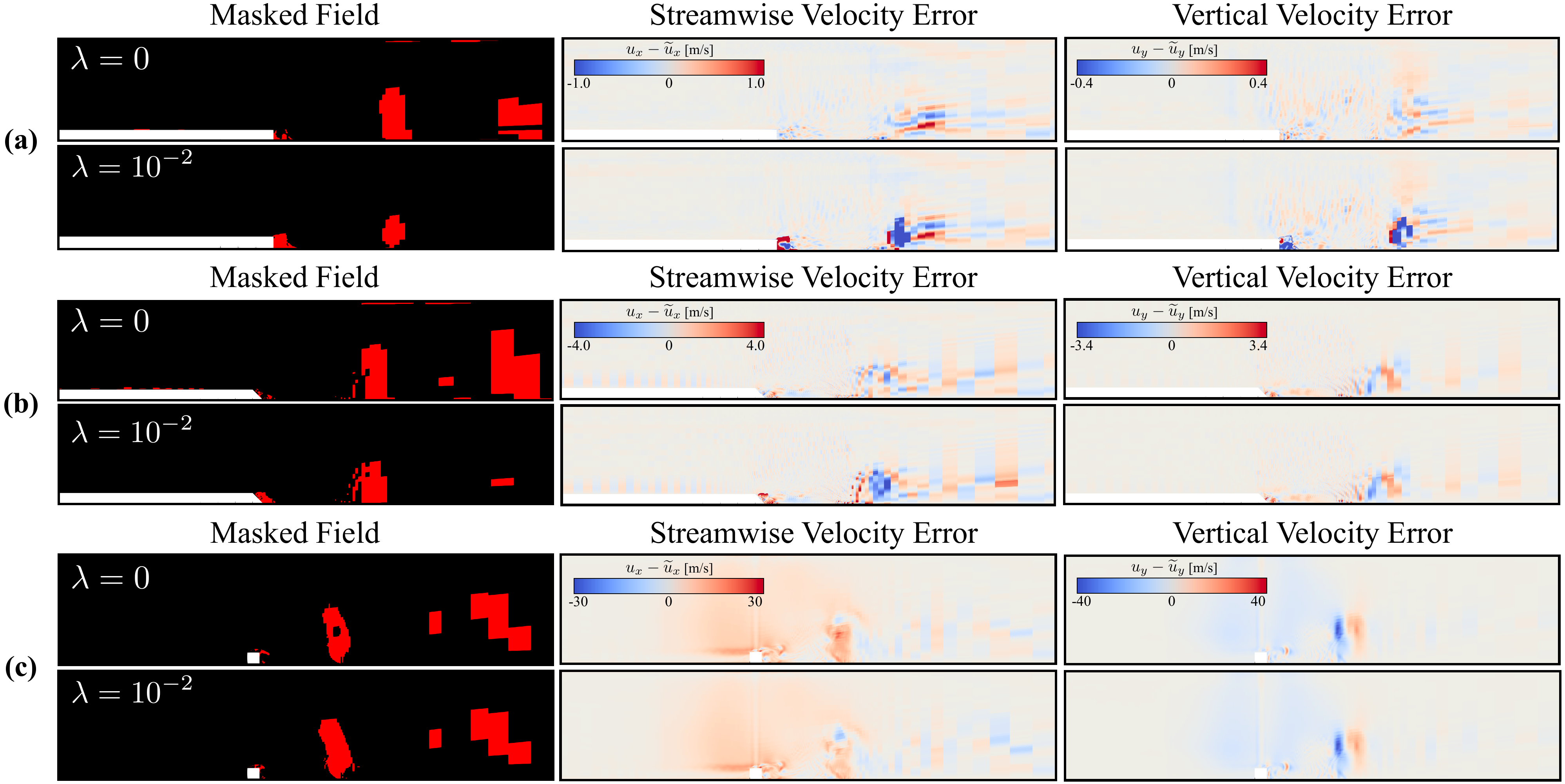}
    \caption{\textbf{(a)} Masked field (left), streamwise velocity error (middle), and vertical velocity error (right) for $\lambda=0$ and $\lambda=10^{-2}$ models. Reproduced from Fig.~\ref{fig:budget_rollout_mask}. \textbf{(b)} Same as (a), but for ramp geometry. \textbf{(c)} Same as (a) and (b), but for wall-mounted cube geometry. Results are shown for $RF=16$ models using inputs from Re=35,392 trajectories.}
    \label{fig:mask_geom_comparison}
\end{figure}

\textbf{Feature Identification and Error Tagging:}
To highlight error correlations, Fig.~\ref{fig:mask_geom_comparison} provides instantaneous masked fields and corresponding spatial distributions of velocity component errors for the three aforementioned geometries: the backward-facing step, the ramp, and the wall-mounted cube. Analogous to Fig.~\ref{fig:budget_rollout_mask}, the figure includes outputs from {interpretable GNN}s trained with different budget regularization factors on the BFS geometry. 

For the BFS configuration in Fig.~\ref{fig:mask_geom_comparison}(a), the {interpretable GNN}s identify coherent structures downstream of the step. A similar identification of coherent structures is seen for the new geometries, in similar parts of the domain relative to the geometric feature. Most notably, the shift in the masked field towards the step anchor with increasing $\lambda$ -- observed originally in BFS configuration -- is also observed in the ramp configuration in Fig.~\ref{fig:mask_geom_comparison}(b). This shift is not seen in the cube case, which shows a markedly lower sensitivity to $\lambda$ than the other two geometries. Despite this, the GNN appears to be identifying coherent structures even in the extrapolated geometries, with the spatial location of these structures generally correlating with high levels of forecasting error. Additionally, the same concentration effect provided by the budget regularization strategy (which is the increase in the amount of error contained in the masked fields with $\lambda$) observed in the BFS case is observed in the ramp geometry -- this effect is again much less apparent in the cube geometry. Although the masked fields are still picking up features in the mounted cube configuration, the lack of $\lambda$-sensitivity in Fig.~\ref{fig:mask_geom_comparison}(c) is likely due to the high absolute errors observed in the respective forecasts. Unlike the BFS and ramp cases, the cube-induced flowfield is not only governed by separation, but also the physical effects of flow impingement upstream of the geometry. Additionally, as described in Appendix~\ref{app:a}, the range in cells resolutions observed in the mesh for the cube configuration is greater than the other two cases, which may also be contributing to the higher errors. Ultimately, however, the extrapolations show how the masked fields largely preserve their feature identification capability in unseen geometries and largely remain correlated with regions of spatially concentrated forecasting error. 

\textbf{Mask Decoherence During Rollout:} Alongside analysis of spatial distributions and error tagging capability, it is of particular interest to evaluate whether the trends in the {interpretable GNN}s \textit{during rollout} also translate to extrapolated geometries. More specifically, recall that through Fig.~\ref{fig:rollout_plots}(d), presence of high-frequency noisy artifacts in the masked fields were shown to be intrinsically tied to the emergence of rollout errors in the BFS configuration. This emergence of noise is referred to as "decoherency". If a similar decoherency is observed during rollout for different geometries, the relationship between such a coherency breakdown and the presence of prediction errors is further verified (i.e., it strengthens confidence in the utility the masked fields provide in terms of predicting the error itself). 

To facilitate such a comparison, it is necessary to produce a unified quantitative measure for the level of coherency/decoherency in the masked fields at a particular point in time, such that a time-series for the coherency measure can be produced for a given model evaluated on different geometries and for different Reynolds numbers. For example, consider the masked fields in Fig.~\ref{fig:mask_geom_comparison}, which are generated after a single-step prediction. They contain structures that are both disjoint and coherent -- an ideal metric should be able to correlate both characteristic qualities to one number, such that emerging "decoherence" can be assessed by monitoring this number in time. 

To this end, the DBSCAN \cite{dbscan} algorithm is executed \textit{in physical space} on the subset of nodes identified in the masked field to produce the desired quantitative metric for coherency, where the metric in the end is the number of DBSCAN clusters. As an idealized example, two DBSCAN clusters encode two "coherent" features in the mask. As noise and decoherency enters the masked field during rollout, the number of DBSCAN clusters is expected to increase -- the rate of increase in the number of clusters therefore encodes the emergence of decoherency in the masked field. 

To summarize, DBSCAN is a popular density-based clustering algorithm that has found use in a wide range of applications. Given a set of points in a feature space, the algorithm returns cluster assignments of these points based on relative distances in this feature space. There are two primary inputs to the standard DBSCAN algorithm: a radius lengthscale $\varepsilon$ that defines the extent to which a neighboring point is reachable from a query point, and a minimum number of interconnected points $n_p$ required to instantiate a cluster. The complete details of the algorithm can be found in Ref.~\cite{dbscan} and are beyond the scope of this work. The DBSCAN implementation in Scikit-learn \cite{pedregosa2011scikit} is used here. 

Before executing the algorithm, the masked fields are first mapped to a cropped structured grid of cell size $\Delta x = L/50$ using nearest-neighbor interpolation, where $L$ is the characteristic lengthscale (e.g., the step, ramp, or cube height). DBSCAN is then executed on the subset of structured grid cells covered by the mask using $\varepsilon = 3\Delta x$ and $n_p=3$. In this setting, $n_p$ is set to be small enough such that de-coherency is identified through a higher number of DBSCAN clusters, but larger than 1 to eliminate potential outliers. 

\begin{figure}
    \centering
    \includegraphics[width=\textwidth]{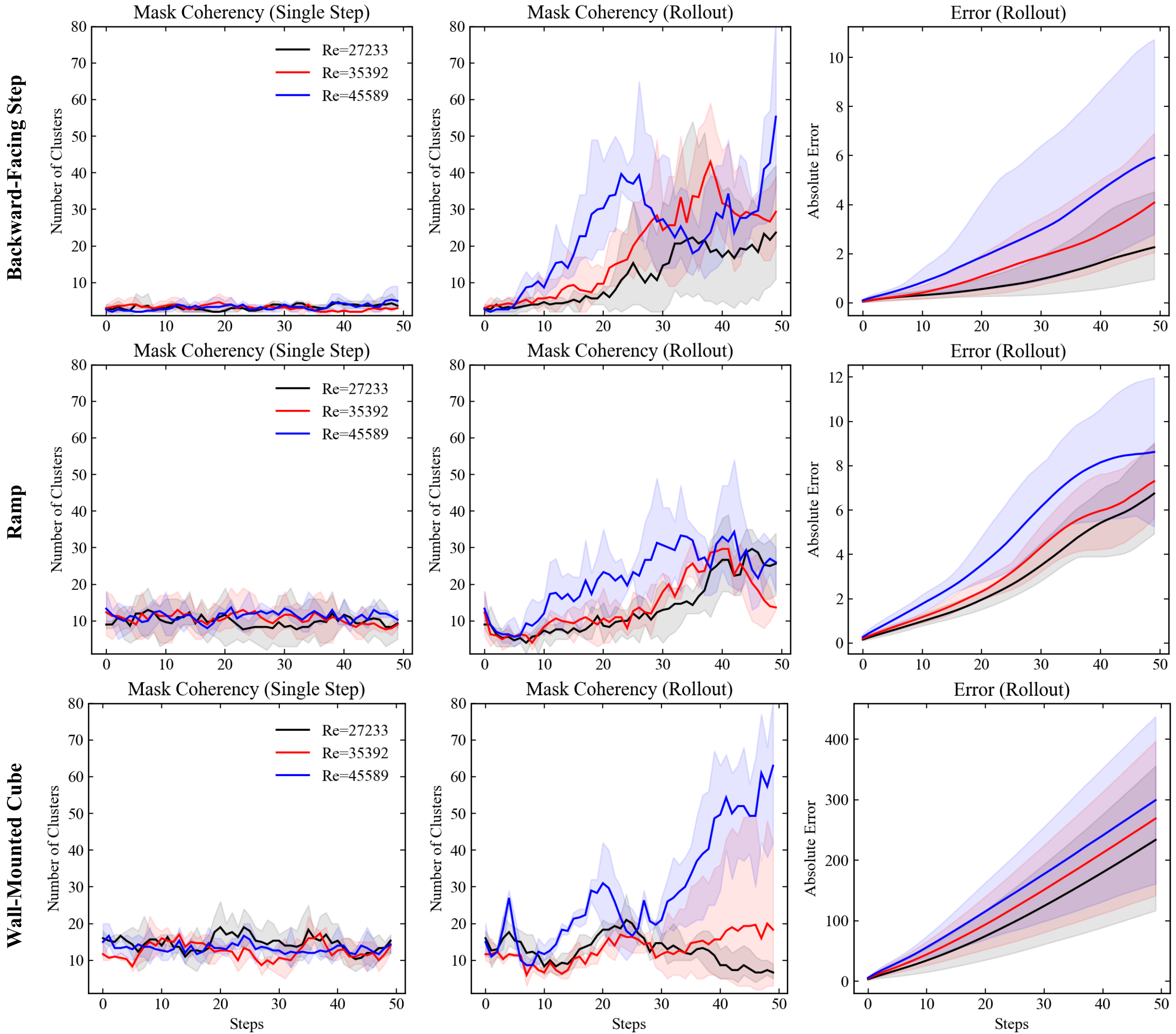}
    \caption{\textbf{Left column:} Mask coherency measure (number of DBSCAN clusters) versus time steps for single-step predictions. \textbf{Middle column:} Mask coherency measure versus time steps for rollout predictions. \textbf{Right column:} Absolute error in velocity field averaged over streamwise and vertical components during rollout versus time steps. Rows, from top to bottom, correspond to BFS, ramp, and cube configurations respectively. Curves shown for initial conditions sourced from Re=27233 (black), 35392 (red), and 45589 (blue) trajectories. Bold lines indicate averages over three initial conditions, with shaded regions providing min-max bounds. Predictions obtained from $\lambda$=0 model with RF=16.}
    \label{fig:mask_coherency}
\end{figure}

The mask coherency in both single-step and rollout settings, juxtaposed with prediction rollout errors, is provided in Fig.~\ref{fig:mask_coherency} for the three geometries at different Reynolds numbers. In the single-step setting (left column of Fig.~\ref{fig:mask_coherency}), the number of clusters is effectively steady. Additionally, the single-step curves are translated up along the y-axis (higher average number of clusters) when moving from the BFS to ramp geometry, and again from the ramp to cube geometry, correlated with the increase in single-step error for these cases observed in Fig.~\ref{fig:mask_geom_comparison}. Interestingly, in the rollout setting, the mask coherency curves are qualitatively similar to the rollout error curves, particularly at early time steps. Most notably, the number of clusters is roughly proportional to the Reynolds number during the early rollout stage, mirroring the trends in rollout error. It should be noted that while rollout errors in the ramp case are quite similar in magnitude to the BFS case, the rollout errors for the cube case are an order of magnitude higher than the BFS and ramp cases. To reiterate, this is likely a consequence of the presence of completely new flow physics by means of impinging flow on a flat surface, an effect not captured in the BFS training data. Despite this, coherency trends in the cube configuration for both single-step and rollout predictions are quite similar to BFS and ramp counterparts.

The trends in Fig.~\ref{fig:mask_coherency} ultimately reveal a useful pathway for leveraging such a coherency measure -- an explicit function of the masked field produced in the GNN forward pass -- as a surrogate for global forecasting error. This in turn opens up promising pathways with regards to producing stabilized forecasting models that are purely data-based.

\section{Conclusion}
This work introduces an {interpretability enhancement} strategy for graph neural network based surrogate models, with focus on unstructured grid based fluid dynamics. More specifically, given a baseline GNN, the strategy consists of (a) appending an interpretable module to the GNN architecture through a skip connection, and (b) an isolated training stage to optimize the parameters introduced by this appended module. The resultant {interpretable GNN} adds interpretability to the baseline GNN by means of a graph pooling layer that acts as an adaptive sub-graph sampler, where these sub-graphs are identified in accordance to the underlying objective function (which here is the mesh-based forecasting error). Through this pooling operation, it is shown how the GNN forward pass generates \textit{masked fields} as a by-product, which isolates regions in physical space in which these sampled sub-graphs are active. The features contained in these masked fields allow one to interpret the relationship between the input flow field at the current time instant, and the model-derived forecast at a future time instant. Through visualization of the masked fields, the {interpretable GNN}s allow one to access regions in physical space intrinsically linked to the modeling task (forecasting), thereby making the model behavior directly accessible from the perspective of the objective function used during training. As a result, the structures identified by pooling operation serve as an accessible link between the forecast-based objective function, the baseline model architecture, and the underlying physics of the problem being modeled. 

Additionally, by adding a regularization term to the objective function during the {interpretability enhancement} procedure in the form of an error budget, it was shown how the mask fields can also be used to identify, during inference, nodes that describe a majority of the anticipated forecasting error. Since the masked fields are an explicit function of the input flowfield (and not the predicted future flowfield), this regularization step adds a form of interpretable error-tagging capability to the modeling framework. A strategy to guide optimal selection of the regularization hyperparameter was also introduced, resulting in a novel approach that adds error-tagging capability to baseline GNNs without sacrificing baseline prediction accuracy. Additional analysis was performed on unseen geometries, with the goal of verifying the identification and error-tagging capability of the masked fields in challenging extrapolation scenarios. The analysis revealed how key GNN properties -- particularly error localization and mask field decoherency -- can indeed translate to unseen geometries.

The authors believe that the strategies introduced here begin to address the limitations of existing black-box mesh-based modeling strategies, and can readily be extended to other applications both within and outside of fluid dynamics surrogate modeling. {It should be emphasized that although the identification of coherent structures is \textit{enabled} by the Top-K sub-sampling mechanism, such identification is ultimately tied down to the physics being modeled (represented by both the domain configuration through initial and boundary conditions, and the objective function used during optimization). Depending on the problem, it is up to the end-user to assign physical meaning to the identified structures. For this reason, the authors believe that a similar coherent identification property of the masked fields is expected to be preserved in other physical applications/configurations, particularly those characterized by dominant instantaneous features that contribute significantly to the dynamics of the system in question (e.g., shock waves, detonation waves, vortices, etc). For more complex turbulent flows in three spatial dimensions and at high Reynolds numbers, it is unclear if the method will produce masked fields that identify (a) many disjoint small patches in physical space, (b) a few disjoint large patches, or (c) some combination of the two; further investigation in this direction is required.}

{To conclude, there are many avenues for future work. The objective functions studied here concentrated on single-step forecasting errors -- an interesting direction for future work is to append the same interpretable sub-sampling module, but include more rollout steps during the training process, thereby producing an interpretable stability enhancement.} Including physics-based loss terms is another promising direction, as it would allow one to investigate the relationship between the identified structures and governing equation residuals. Additionally, although the modeling context in this work focused on a node-level regression task, extension of the ideas presented here for graph-level (i.e., more macroscopic) regression tasks may be possible. Regarding treatment of multiple geometries, a promising direction is to {enhance} a GNN for interpretability on geometry B using a baseline trained on geometry A. {Further, so long as the input is represented as a graph, the authors believe that the interpretability and error indication enhancements developed here can be readily extended to different graph-based (and more complex sequence-based) architectures, such as graph transformers.} {Lastly, from the feature identification angle, establishing a connection between the Top-K procedure used here and other unsupervised feature extraction approaches (based on data-driven balance models \cite{callaham2021learning} and physics-informed clustering \cite{jskmeans}, for example) is an interesting direction.} These aspects, as well as other directions, are left for future studies.

\section{Acknowledgements} 
The manuscript has been created by UChicago Argonne, LLC, Operator of Argonne National Laboratory (Argonne). The U.S. Government retains for itself, and others acting on its behalf, a paid-up nonexclusive, irrevocable world-wide license in said article to reproduce, prepare derivative works, distribute copies to the public, and perform publicly and display publicly, by or on behalf of the Government. This work was supported by the U.S. Department of Energy (DOE), Office of Science under contract DE-AC02-06CH11357. This research used resources of the Argonne Leadership Computing Facility, which is a U.S. Department of Energy Office of Science User Facility operated under contract DE-AC02-06CH11357. RM acknowledges funding support from ASCR for DOE-FOA-2493 ``Data-intensive scientific machine learning'' and from the RAPIDS2 Scientific Discovery through
Advanced Computing (SciDAC) program.

\bibliography{references}

\newpage

\begin{appendices}

\section{Graph Definitions and Backward Facing Step Flow}
\label{app:a}
A graph $G$ is defined as $G = (V, E)$. Here, $V$ is the set of graph vertices or nodes; its cardinality $|V|$ denotes the total number of nodes in the graph. More specifically, each element in $V$ is an integer corresponding to the node identifier, such that $V = \{1, \ldots, |V|\}$. The set $E = \{(i,j)  | i,j \in V \}$ contains the graph edges. A single edge in $E$ is described by two components: a sender node $i$ and a receiver node $j$. The existence of an edge between two nodes signifies a sense of similarity or correlation between these nodes. The set $E$ is equivalent to the graph adjacency matrix, which is a critical input to graph neural network (GNN) based models; it is the mechanism that (a) informs how information is exchanged throughout node neighborhoods in the graph, and (b) enables GNNs to be invariant to permutations in node ordering. Alongside the graph connectivity contained in the edge indices $E$, data on the graph is represented in the form of attribute or feature matrices for nodes and edges. As described in the main text in Sec.~\ref{sec:modeling_task}, the graph $G$ is instantiated directly from the dual of the backward-facing step (BFS) mesh, where nodes coincide with cell centroids and edges with shared face intersections. 

\begin{figure}
    \centering
    \includegraphics[width=\textwidth]{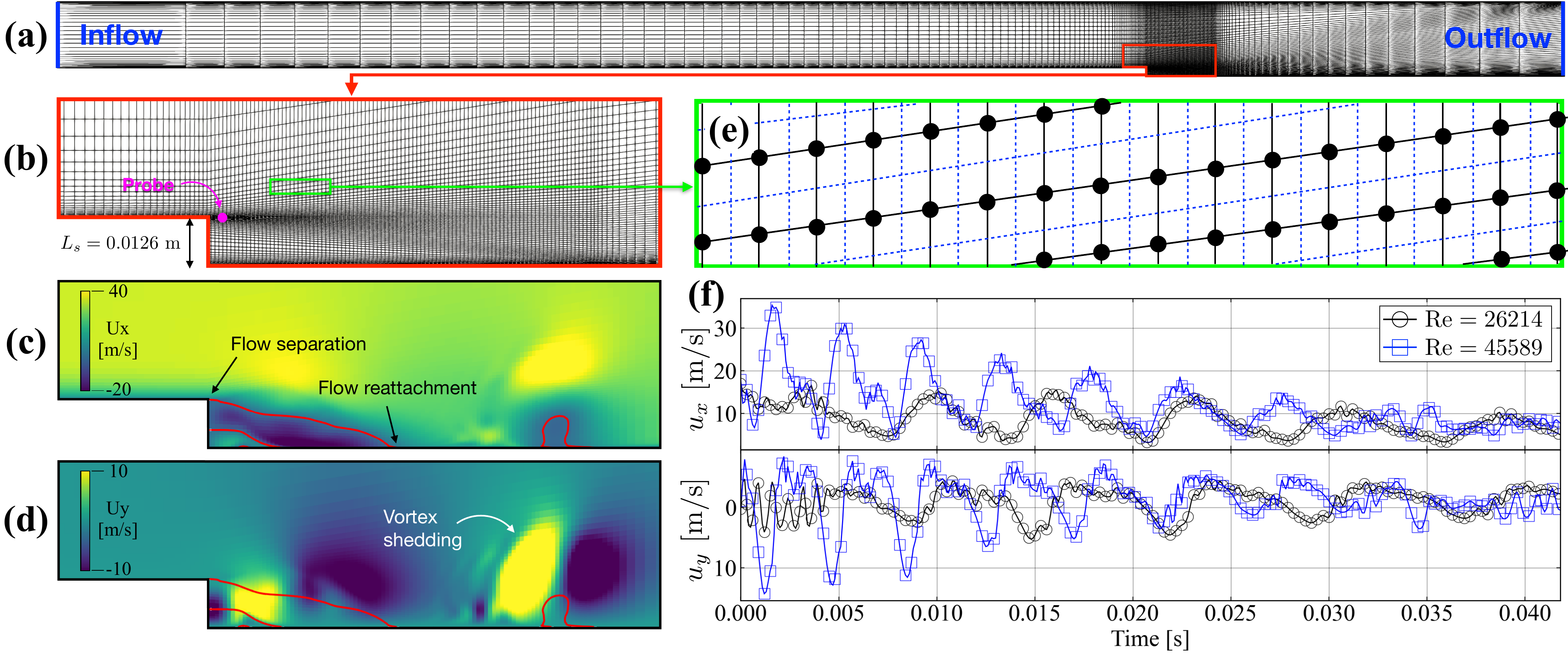}
    \caption{\textbf{(a)} Full mesh of BFS configuration. An inflow boundary condition is provided at the left end, and an outflow at the right. Remaining boundaries are no-slip walls. \textbf{(b)} Zoom-in of the mesh in the near-step region, highlighting the variation in cell size and refinement zones. \textbf{(c)} Instantaneous snapshot of streamwise component of mesh-based velocity field ($U_x$) from the $Re=26,214$ trajectory. Red contour marks locations where streamwise velocity is zero. \textbf{(d)} Same as (c), but for vertical velocity component ($U_y$). \textbf{(e)} Subset of the finite-volume based graph, which is the dual of the mesh. Black markers are nodes (cell centroids), black lines are edges, and blue dashed lines indicate the underlying mesh cells. \textbf{(f)} Streamwise (top) and vertical (bottom) velocity component evolution at the near-step probe location indicated in (b). Black circles denote Re=26214, blue squares denote Re=45589.}
    \label{fig:bfs_mesh}
\end{figure}

Figure~\ref{fig:bfs_mesh}(a) shows the full-geometry BFS mesh, which is the discretized domain of interest. In the BFS configuration, flow enters from an inlet on the left and propagates through an initial channel of fixed width upon encountering a step anchor, triggering flow separation. A zoom-in of the mesh near the step -- which is the key geometric feature -- is shown in Fig.~\ref{fig:bfs_mesh}(b). This geometric configuration is relevant to many engineering applications and admits complex unsteady flow features in the high Reynolds number regime, where the Reynolds number (Re) is given by $\text{Re} = {U L_s}/{\nu}$. Here, $U$ is the inflow (freestream) velocity magnitude, $L_s$ is the height of the backward-facing step, and $\nu$ is the fluid viscosity. 

Each simulation generates a high-dimensional trajectory in time, which is represented as a set of successive instantaneous mesh-based flow snapshots sampled at a fixed time step interval. Figs.~\ref{fig:bfs_mesh}(c) and (d) show the streamwise and vertical velocity component fields composing one such instantaneous snapshot, sourced from an Re=26,214 simulation. Indicated in the figures are the three characteristic flow features of the BFS geometry: (1) flow separation upon encountering the step, (2) flow re-attachment at the bottom-most wall, and (3) vortex shedding induced by the emergence of recirculation zones in the near-step region. These interacting features emerge from the high degree of nonlinearity in the governing equations at the Reynolds numbers considered here. Since the flow is unsteady, the spatial locations of each of these features (i.e., the flow reattachment point) change as a function of time in accordance to a characteristic vortex shedding cycle, which itself is dependent on the Reynolds number. This unsteadiness and Reynolds number dependence is highlighted in Fig.~\ref{fig:bfs_mesh}(f), which showcases the temporal evolution of the flow at a single spatial location in the mesh for two Reynolds numbers.

A visualization of the graph is provided in Fig.~\ref{fig:bfs_mesh}(e), showing the BFS mesh overlaid on top of a subset of the nodes and edges. Combined with the connectivity $E$, input node attribute matrices used for GNN training are recovered directly from the velocity field stored on cell centroids. Since the flow at a particular Reynolds number is unsteady, these fluid velocities are time-evolving, so the features stored in the node attribute matrix can also be considered time-evolving. At a particular time step $m$ along the trajectory, the node attribute matrix is given by
\begin{equation}
    \label{eq:node_attribute}
    {\bf X}_m = 
    \begin{bmatrix}
        u(x_1, m\Delta t) & v(x_1, m\Delta t) \\
        u(x_2, m\Delta t) & v(x_2, m\Delta t) \\
        \vdots & \vdots &  \\
        u(x_{|V|}, m\Delta t) & v(x_{|V|}, m\Delta t)
    \end{bmatrix},   
    \quad m = 1,\ldots,T. 
\end{equation}
In Eq.~\ref{eq:node_attribute}, $u$ and $v$ denote streamwise and vertical velocity components respectively, $x_i$ ($i = 1,\ldots,|V|$) is the physical space location for node $i$, $m$ denotes the temporal index, $\Delta t$ is the trajectory time step, and $T$ is the total number of flow snapshots in the trajectory. A trajectory is defined as 
\begin{equation}
    \label{eq:trajectory}
    \mathcal{T}_n = ( {\bf X}_1, {\bf X}_2, \ldots, {\bf X}_T )_n, \quad n = 1,\ldots,20, 
\end{equation}
where $n$ denotes the trajectory index corresponding to a particular Reynolds number. As discussed in Sec.~\ref{sec:modeling_task}, there are a total of 20 BFS trajectories with 10 allocated for training and the remaining 10 for testing. 

\begin{figure}
    \centering
    \includegraphics[width=\textwidth]{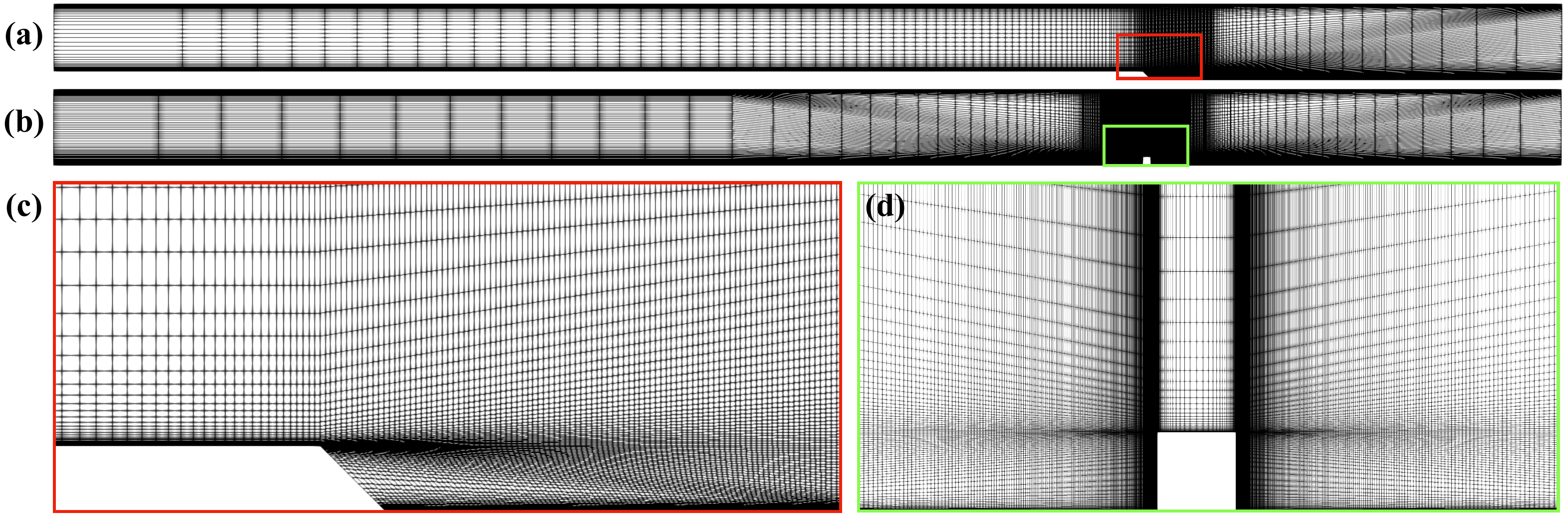}
    \caption{\textbf{(a)} Image of ramp geometry mesh. \textbf{(b)} Image of wall-mounted cube geometry mesh. \textbf{(c)} Zoom-in of ramp mesh, with inset indicated in (a). \textbf{(d)} Zoom-in of cube mesh, with inset indicated in (b).}
    \label{fig:ramp_cube_geometry}
\end{figure}

Analogous to the BFS mesh diagrams shown in Fig.~\ref{fig:bfs_mesh}(a) and (b), meshes for the ramp and wall-mounted cube geometries used throughout Sec.~\ref{sec:results:geom_extrap} are shown in Fig.~\ref{fig:ramp_cube_geometry}. It should be noted that while the total number of cells (as well as range in cell resolutions) is effectively equal between the ramp and BFS cases, the same is not true for the cube case. As indicated by Fig.~\ref{fig:ramp_cube_geometry}, the wall-mounted cube mesh contains overall finer cell resolutions near the geometry, leading to a larger total number of cells (and therefore graph nodes). The mesh refinement near the cube, particularly at the corners, was required to eliminate spurious artifacts in the ground-truth simulations. This, however, results in a high level of resolution change near the cube walls. Is is likely that such levels of mesh resolution transition (greater than those observed in the BFS and ramp cases) also plays a role in generating higher errors for extrapolated predictions on this geometry relative to the other two, as described in Sec.~\ref{sec:results:geom_extrap}.

\section{Data Generation with OpenFOAM}
\label{app:b}

\begin{figure}
    \centering
    \includegraphics[width=0.7\columnwidth]{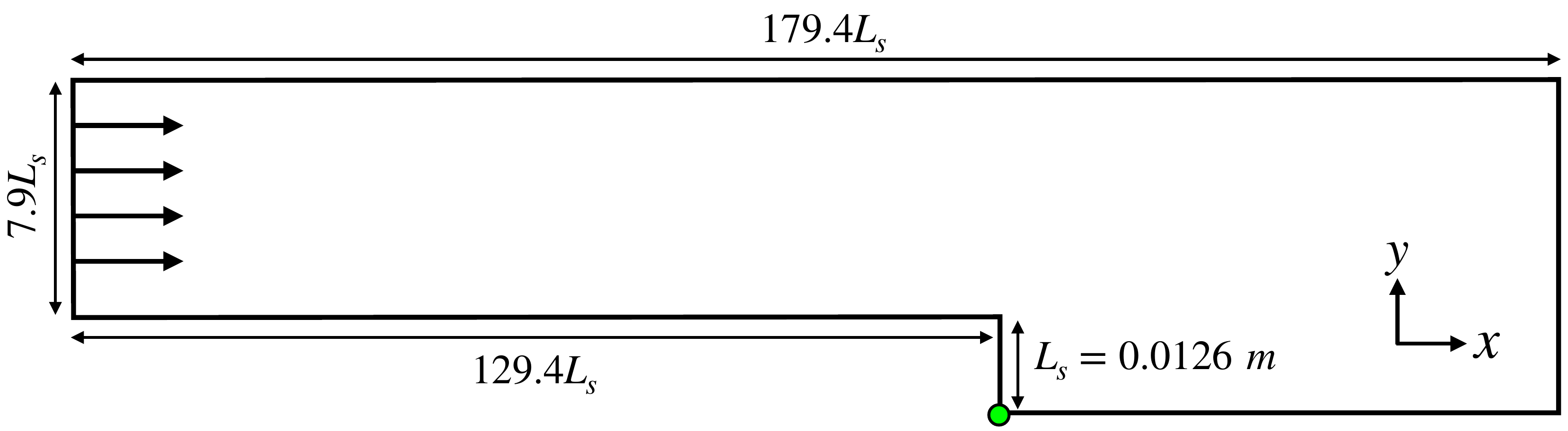}
    \caption{Schematic of BFS geometry (not to scale). Green point indicates grid origin. See Fig.~\ref{fig:bfs_mesh} for full mesh to scale.}
    \label{fig:bfs_geometry}
\end{figure}

Data was generated using computational fluid dynamics (CFD) simulations of the BFS configuration. Additional geometric detail of the BFS configuration is provided in Fig.~\ref{fig:bfs_geometry}. 

The simulations were conducted using \verb|OpenFOAM|, which is an open-source CFD backend that includes a suite of default fluid flow solvers tailored to various physics applications. \verb|OpenFOAM| leverages finite-volume discretization using unstructured grid representations. For this work, the \verb|PimpleFOAM| solver was used, which implements a globally second-order discretization of the unsteady incompressible Navier-Stokes equations \cite{openfoam_fv,piso_1986,simple}. This software is used to solve a filtered version of the two-dimensional Navier-Stokes equations (analogous to large-eddy simulation), written as 
\begin{equation}
    \label{eq:gov_eq}
    \begin{split}
    \frac{\partial \widetilde{u}_j}{\partial t} + \frac{\partial \widetilde{u}_i \widetilde{u}_j}{\partial x_i} &= \nu \frac{\partial^2 \widetilde{u}_j}{\partial x_i \partial x_i} - \frac{\partial \tau_{ij}^r}{\partial x_i} - \frac{1}{\rho}\frac{\partial \widetilde{p}}{\partial x_j}, \\
    \frac{\partial \widetilde{u}_k}{\partial x_k} &= 0.
    \end{split}
\end{equation}
In Eq.~\ref{eq:gov_eq}, $\widetilde{.}$ denotes a spatial (implicit) filtering operation, ${u}_j$ is the time-evolving $j$-th component of velocity, $p$ is the pressure, $\nu$ is a constant kinematic viscosity, $\rho$ is a constant fluid density, and $\tau^r$ is the deviatoric component of the residual, or sub-grid scale (SGS), stress tensor. This is represented here using a standard Smagorinsky model \cite{smagorinsky}, which casts the residual stress $\tau_{ij}^r$ as a quantity proportional to the filtered rate-of-strain as
\begin{equation}
    \label{eq:smagorinsky}
    \begin{split}
    \tau_{ij}^r &= -2 \nu_r \widetilde{S}_{ij},\\
    \nu_r &= \left(C_S \Delta \right)^2 |S|.  
    \end{split}
\end{equation}
In Eq.~\ref{eq:smagorinsky}, $S_{ij}$ is the filtered rate-of-strain, $|S| = \sqrt{2S_{ij}S_{ij}}$ is its magnitude, $\nu_r$ is the turbulent eddy-viscosity, $C_S$ is the Smagorinsky constant ($C_S=0.168$ here), and $\Delta$ is the box-filter width. These simulation parameters, alongside the mesh and boundary condition specifications required to run the simulation, are provided through input files in the \verb|OpenFOAM| case directory.

{As described in the beginning of Sec.~\ref{sec:modeling_task}, using the \verb|OpenFOAM| solver, training and testing data was generated in the step height Reynolds number range of [26214, 45589]. As a reference, for this step height Reynolds number range of [26214, 45589], the corresponding friction Reynolds number range was found to be [3296, 4986]. To compute the friction Reynolds number, a characteristic length scale of half the initial channel height was used (half the channel height before encountering the step). The characteristic velocity was computed by taking a space-time average of the friction velocity using quantities at the lower wall upstream of the step anchor; space averages were taken in the streamwise (x) coordinate range of [$-5L_s$, $0$], where $L_s$ is the step height and 0 is the x-coordinate location of the step anchor. Friction velocities were computed with the help of the \verb|wallShearStress| post-processing function in \verb|OpenFOAM|.}

\section{Multiscale Message Passing Layer}
\label{app:c}

As illustrated in Fig.~\ref{fig:arch}, the baseline GNN architecture and the {interpretable sub-sampling module} leverage multiscale message passing (MMP) layers. In short, MMP layers invoke a series of standard message passing layers on a hierarchy of graphs corresponding to different characteristic lengthscales. This section briefly outlines the formulation of these layers. Before proceeding, it should be noted that the general multi-scale message passing concept, inspired by methods used in multigrid algorithms for solving partial differential equations, has been successfully applied in several mesh-based modeling works \cite{multiscale_meshgraphnets,lino_gnn,shivam_topk,deshpande2024magnet}. 

\begin{figure}
    \centering
    \includegraphics[width=\textwidth]{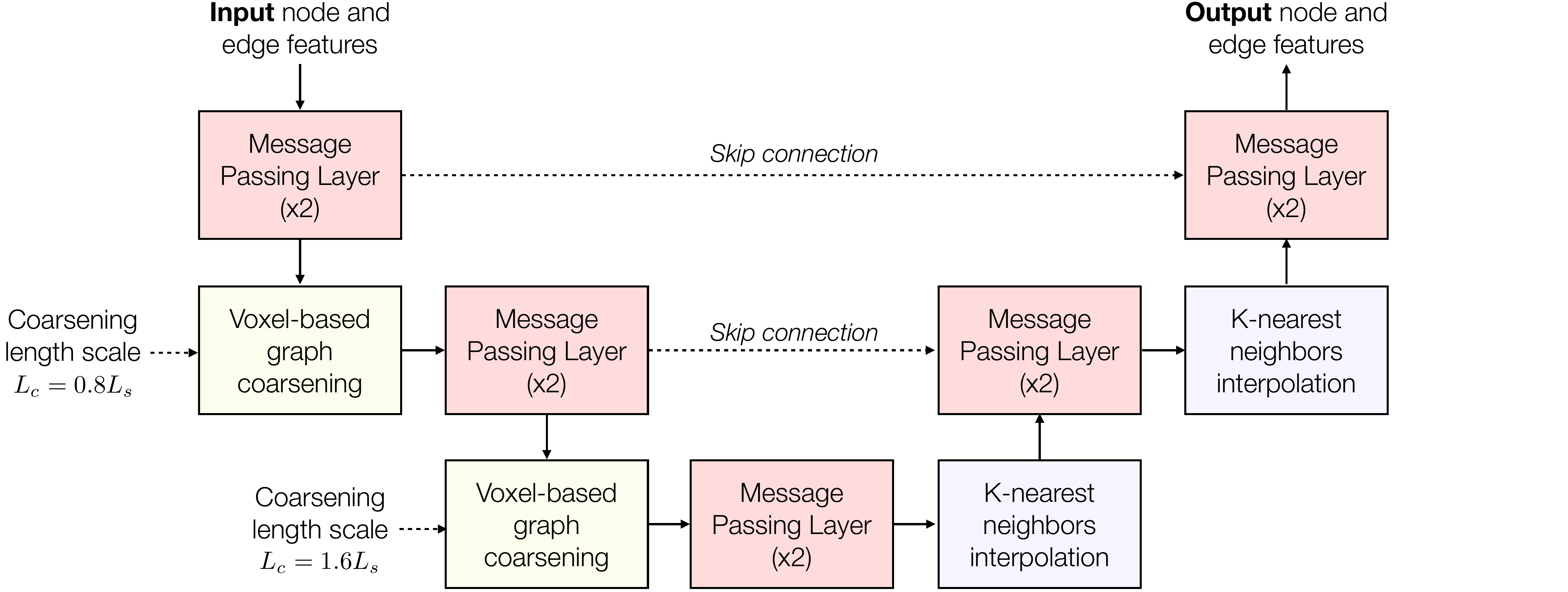}
    \caption{Multiscale message passing (MMP) layer used in baseline and {interpretable GNN} architectures in Fig.~\ref{fig:arch}.}
    \label{fig:mmp_layer}
\end{figure}

A schematic of the MMP layer architecture used in this work is shown in Fig.~\ref{fig:mmp_layer}, and follows the general formulation of the same layer used in recent work \cite{shivam_topk}. Note that a single MMP layer in this context can be interpreted as a U-net type architecture. There are three components: (1) single-scale message passing, (2) a graph coarsening operation, and (3) a graph interpolation operation.

\underline{Message passing:} A message passing layer updates node and edge features without modifying the graph connectivity. One single-scale message passing layer consists of the following operations:
\begin{align}
  \text{Edge update: } {\bf e}_{ij}^p &= {\cal F}_e ({\bf e}_{ij}^{p-1}, {\bf x}_i^{p-1}, {\bf x}_j^{p-1}) \in \mathbb{R}^{N_H}\\
  \text{Edge aggregation: } {\bf a}_{i}^p &= \frac{1}{|\mathcal{N}(i)|} \sum_{j \in \mathcal{N}(i)} {\bf e}_{ij}^p \in \mathbb{R}^{N_H}\\
  \text{Node update: } {\bf x}_{i}^p &= {\cal F}_n ({\bf a}_{i}^p, {\bf x}_i^{p-1}) \in \mathbb{R}^{N_H}. 
\end{align}
In the above equations, the superscript $p$ denotes the message passing layer index, ${\bf e}_{ij}$ is the edge feature vector corresponding to sender and receiver node indices $i$ and $j$ respectively, ${\bf x}_i$ is the feature vector for node $i$,  $\mathcal{N}(i)$ is the set of neighboring node indices for node $i$, and ${\cal F}_e$ and ${\cal F}_n$ are independently parameterized multi-layer perceptrons (MLPs). The quantity ${\bf a}_{i}$ represents the neighborhood-aggregated edge features corresponding to node $i$, and $N_H$ is the hidden feature dimensionality (assumed to be the same for nodes and edges after the action of the encoder).

\underline{Coarsening and Interpolation}: As the name implies, the goal of graph coarsening in this context is to coarsen a given input graph such that the average edge lengthscale (in terms of relative distance in physical space) is increased. There are a number of ways to achieve graph coarsening -- here, a voxel clustering strategy is used, which produces the coarse graph using voxelization based on an input coarsening lengthscale. Briefly, in this approach, a graph is coarsened by first overlaying a voxel grid, where the centroids of the voxel cells coincide with the coarsened graph nodes. The underlying fine graph nodes can then be assigned to the voxel cells/clusters via computation of nearest centroids, where the coarse node features are initialized using the average of fine node features within a cell. Edges between coarse nodes are added if fine graph edges intersect a shared voxel cell face -- the coarse graph edge features are then initialized by averaging fine edge features for edges that satisfy this intersection. For interpolation, a K-nearest neighbors (KNN) strategy is used which amounts to linear interpolation to populate fine node features with a user-specified K number of nearby coarse nodes (K=4 in the MMP layers used in this work). Voxelization and KNN interpolation are implemented using the \verb|voxel_grid| and \verb|knn_interpolate| functions available in PyTorch Geometric \cite{pytorch_geom}.

\section{Architecture and Training Details}
\label{app:d}

All models in this study were implemented using PyTorch \cite{pytorch} and PyTorch Geometric \cite{pytorch_geom} libraries. Following Fig.~\ref{fig:arch}, a total of 2 MMP layers were utilized in the processor (L=2), the encoder and decoder operations leverage three-layer MLPs, and all message passing layers leverage two-layer MLPs. All MLPs incorporate exponential linear unit (ELU) activation functions with residual connections. Layer normalization \cite{layernorm} is used after every MLP evaluation. The hidden feature dimensionality is set to $N_H=128$.

Input node features in ${\bf X}_m$ correspond to streamwise and vertical velocity components, target node features are the same features at a future time step ${\bf X}_{m+1}$, and edge features in ${\bf E}$ are initialized using relative physical space distance vectors between the respective nodes. Before training, the data was standardized using training data statistics, and 10\% of training data was set aside for validation purposes. As per previously reported findings \cite{meshgraphnets}, a small amount of noise was injected into the input mesh-based flowfields for stability purposes; each input node feature was perturbed independently by scalar value sampled from a normal distribution with standard deviation of $10^{-2}$ before training. 

The Adam optimizer \cite{adam} equipped with a plateau-based scheduler (the PyTorch function \verb|ReduceLROnPlateau|) was utilized during training with an initial learning rate of $10^{-5}$ and minimium learning rate fixed to $10^{-7}$. The batch size used during training was $8$. Each model was trained using 8 Nvidia A100 GPUs hosted on two nodes of the Polaris supercomputer at the Argonne Leadship Computing Facility.

{\textbf{Parameter counts and computational costs:}  The baseline model (which includes node and edge encoders, node decoder, and 2 MMP layers -- refer to Fig.~\ref{fig:arch}) contains $2.41\times 10^{6}$ parameters. The {interpretable sub-sampling module} parameters are populated by a single Top-K layer and 1 MMP layer, producing $1.16\times 10^{6}$ parameters and a value of 0.48 for the ratio (\# {interpretable sub-sampling module} parameters / \# baseline model parameters), which is fixed across all experiments. It should be noted that in the {interpretable module}, practically all parameters stem from the MMP layer, with only $N_F=128$ (the hidden graph node feature dimensionality) parameters contained in the Top-K layer (the projection vector); as such, this ratio can be readily reduced by specifying a lower-complexity MMP layer in the {interpretable sub-sampling module}. 

Training times will ultimately depend on hardware, batch sizes, and dataset/graph sizes. However, for reference, using 8 Nvidia A100 GPUs, a training set size of roughly 40k snapshots, and a graph batch size of 2 (per GPU), the training time for a single epoch using the baseline GNN was 106.4 seconds, and the corresponding training time for the {interpretable GNN} was 90.2 seconds. Although the model size itself is larger, the lower training time in the {interpretable GNN} comes from the fact that all baseline model parameters are frozen (rendering a less intensive back-propagation). On the other hand, in the inference stage, it was found that the cost of {interpretable GNN}s were roughly 2x the cost of the baseline (roughly the same proportion as the parameter ratio).  
}

{
\section{Masked Field Sensitivity and Stability}
\label{app:e}

    The \verb|tanh| activation function was used in the Top-K layer (refer to Fig.~\ref{fig:topk}) as it is the recommended/default setting in the PyTorch Geometric implementation \cite{pytorch_geom}. However, it is important to understand the sensitivity of the masked fields to the activation function employed during the pooling stage. To shed some light on this front, this section provides additional numerical experiments that replace the \verb|tanh| activation in the Top-K layer with a \verb|relu| activation, with the goal of understanding the effect of this change on the identified masked fields. 
    
    Figure~\ref{fig:revision:1} shows a comparison of masked fields generated from single-step predictions at two different time instances using these activation settings (to facilitate this comparison, {interpretable GNN}s were trained using at the same Top-K reduction factor of RF=16 without any budget regularization, i.e., $\lambda=0$). The figure highlights how the identified regions are indeed sensitive to the activation function used, in that the \verb|tanh| model (used throughout the main text) picks up additional structures in the far-downstream region of the step that are not identified by the \verb|relu| model. However, despite this, the key structures near the step and in the near-downstream region are picked up by both models, and are qualitatively similar. 

    It should be noted that, although not shown here, the GNN leveraging the \verb|relu| activation in the Top-K layer resulted in a higher overall loss (worse performance) than the \verb|tanh| counterpart, which is likely the major contributor to the differences in the masked fields in Fig.~\ref{fig:revision:1}. A similar trend is observed in Fig.~\ref{fig:budget_rollout_mask} in the main text: the models with higher values of $\lambda$, when compared to the $\lambda=0$ model, also produce a higher MSE loss and do not pick up the far-downstream structures. 
    
    Ultimately, Fig.~\ref{fig:revision:1} suggests that changing the activation function in the Top-K layer does not fundamentally alter the model's ability to pick up key coherent structures in the masks, though the authors recommended using \verb|tanh| in the Top-K layer for these applications. It is hypothesized that the bounded nature of the \verb|tanh| function helps in the training process for the Top-K vector, but a more comprehensive study on the effect of activation functions is warranted, as with any neural network based model (there are many more candidate functions to choose from). 

    \begin{figure}
        \centering
        \includegraphics[width=\textwidth]{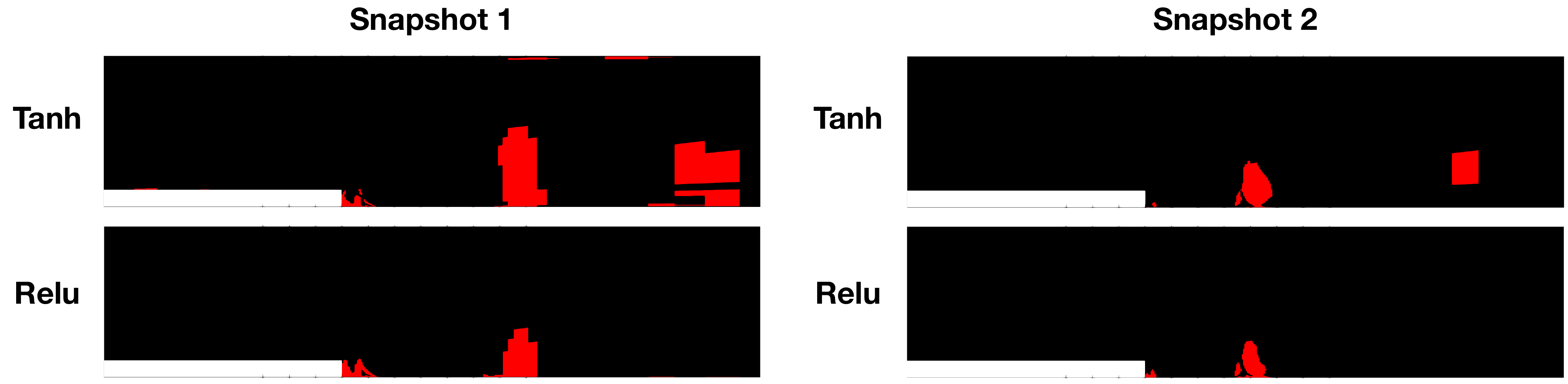}
        \caption{{\textbf{(Left)} Masked field visualizations comparing {interpretable GNN} trained using tanh activation in the Top-K layer (top) and relu activation in the Top-K layer (bottom). \textbf{(Right)} Same as left, but for a different input velocity field (different snapshot). Both snapshots are extracted at Re=35392 (trends were found to be similar at all other tested Reynolds numbers).}}
        \label{fig:revision:1}
    \end{figure}

    The above discussion is centered around characterizing the stability of the masked field in response to architectural changes. An alternative viewpoint (that also serves to verify the physical significance of the identified structures) is to extract the masked field sensitivity to parameter perturbations in the \textit{same} GNN architecture. To this end, Fig.~\ref{fig:revision:2} shows spatial distributions of the standard deviations of masked fields produced using an ensemble of 30 different {interpretable GNN} models of the same architecture (i.e., each member in the ensemble has the same architecture as used in the main text, with \verb|tanh| activations in the Top-K layer). The model ensemble was constructed by first taking a converged {interpretable GNN} model (RF=16, $\lambda=0$), and then performing stochastic gradient descent (SGD) training iterations on the converged model with a small learning rate of $10^{-7}$. The model ensemble was then populated by saving the parameter states at each of these SGD iterations (this strategy has been used in previous work for uncertainty quantification of machine learning models \cite{morimoto2022assessments}). The result is a set of models with the same effective performace/loss, but ``perturbed" parameter states, offering a useful testbed for studying the stability/consistency of structures in the masked fields through statistical analysis.

    \begin{figure}
        \centering
        \includegraphics[width=\textwidth]{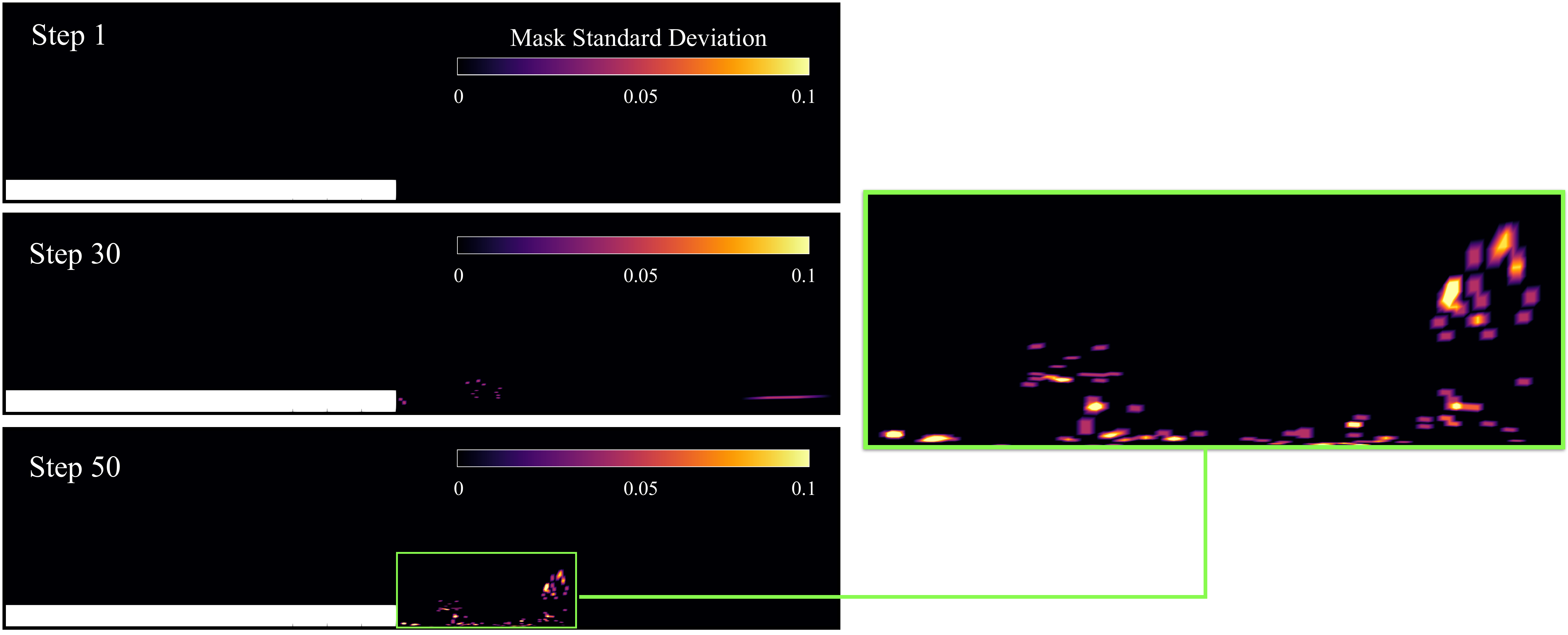}
        \caption{{Masked field standard deviations during rollout steps 1, 30, and 50 at Re=35392. Standard deviations produced using RF=16 {interpretable GNN}s with $\lambda=0$.}}
        \label{fig:revision:2}
    \end{figure}

    The plots in Fig.~\ref{fig:revision:2} are computed by feeding the same input to each of the models in the ensemble, collecting the respective masks (there will be 30 masks at each time instant, since there are 30 models in the ensemble), and then computing the node-wise standard deviation in the masks. Figure~\ref{fig:revision:2} shows these standard deviations at three different rollout steps (1, 30, and 50) using an initial condition from the Re=35392 trajectory. Interestingly, the figures imply that the masks produced by the Top-K sub-sampling are relatively stable until about 30 steps, with noticeable standard deviations appearing at 50 steps and beyond concentrated around the near-step region. The emergence of the mask standard deviation at around 25-30 steps correlates with the high rate of growth in the masked field decoherence shown in Fig. 12 in the original manuscript near roughly the same point. Ultimately, the results in Fig.~\ref{fig:revision:2} serve to verify stability in the masked field structures in the early rollout stages. Additionally, this initial assessment reveals a promising direction in the form of connecting masked field decoherence in a single model to ensemble-based model uncertainties. A more detailed exploration of this connection is left for future studies. 
}

\end{appendices}

\end{document}